\documentclass{article}

    \PassOptionsToPackage{numbers, compress}{natbib}
\newcommand{\website}{https://sites.google.com/view/simone-scene-understanding/}
    \usepackage[final]{neurips_2021}

\usepackage[utf8]{inputenc} % allow utf-8 input
\usepackage[T1]{fontenc}    % use 8-bit T1 fonts
\usepackage{hyperref}       % hyperlinks
\usepackage{url}            % simple URL typesetting
\usepackage{booktabs}       % professional-quality tables
\usepackage{amsfonts}       % blackboard math symbols
\usepackage{nicefrac}       % compact symbols for 1/2, etc.
\usepackage{microtype}      % microtypography
\usepackage{xcolor}         % colors
\usepackage{multirow}
\usepackage{bm}
\usepackage{amsmath}
\usepackage{amssymb}
\usepackage{bbold}
\usepackage{mathtools}
\usepackage{subcaption}
\usepackage{hyperref}
\usepackage{xcolor}
\usepackage{ifthen}
\usepackage[inline]{enumitem}

\usepackage{array}
\newcolumntype{P}[1]{>{\centering\arraybackslash}p{#1}}

\definecolor{medium-blue}{rgb}{0,0,1}

\hypersetup{colorlinks, urlcolor={medium-blue}}

\newcommand{\obj}{\mathbf{o}}
\newcommand{\view}{\mathbf{f}}
\newcommand{\loc}{\mathbf{l}}
\newcommand{\mask}{m}
\newcommand{\x}{\mathbf{x}}
\newcommand{\X}{\mathbf{X}}
\newcommand{\mean}{\bm{\mu}}
\newcommand{\decoder}{\mathcal{D_{\theta}}}
\newcommand{\encoder}{\mathcal{E_{\phi}}}
\newcommand{\simonet}{SIMONe}
\DeclarePairedDelimiterX{\infdivx}[2]{(}{)}{%
  #1\;\delimsize\|\;#2%
}
\newcommand{\kldiv}{D_{KL}\infdivx}

\newcommand{\expnumber}[2]{{#1}\mathrm{e}{#2}}

\title{\simonet{}: View-Invariant, Temporally-Abstracted Object Representations via Unsupervised Video Decomposition}

\author{%
  Rishabh~Kabra$^1$, Daniel~Zoran$^1$, Goker~Erdogan$^1$, Loic~Matthey$^1$ \\
  \textbf{Antonia~Creswell$^1$, Matthew~Botvinick$^1$, Alexander~Lerchner$^1$, Christopher~P.~Burgess$^{2*}$}\\
  $^1$DeepMind, $^2$Wayve, $^*$Work done at DeepMind\\
  \texttt{\{rkabra, danielzoran, gokererdogan, lmatthey,}\\
  \texttt{tonicreswell, botvinick, lerchner\}@deepmind.com, chrisburgess@wayve.ai} \\
}

\begin{document}

\maketitle

%%%
%\vspace{-10pt}
\begin{abstract}
    To help agents reason about scenes in terms of their building blocks, we wish to extract the compositional structure of any given scene (in particular, the configuration and characteristics of objects comprising the scene). This problem is especially difficult when scene structure needs to be inferred while also estimating the agent’s location/viewpoint, as the two variables jointly give rise to the agent’s observations. We present an unsupervised variational approach to this problem. Leveraging the shared structure that exists across different scenes, our model learns to infer two sets of latent representations from RGB video input: a set of "object" latents, corresponding to the time-invariant, object-level contents of the scene, as well as a set of "frame" latents, corresponding to global time-varying elements such as viewpoint. This factorization of latents allows our model, \simonet{}, to represent object attributes in an allocentric manner which does not depend on viewpoint. Moreover, it allows us to disentangle object dynamics and summarize their trajectories as time-abstracted, view-invariant, per-object properties. We demonstrate these capabilities, as well as the model's performance in terms of view synthesis and instance segmentation, across three procedurally generated video datasets.     
\end{abstract}

%\vspace{-10pt}
\section{Introduction}
\label{sec:intro}

The problem of \emph{unsupervised visual scene understanding} has become an increasingly central topic in machine learning \cite{malik_2015,inbooksceneunderstanding}. The attention is merited by potential gains to reasoning, autonomous navigation, and myriad tasks. However, within the current literature, different studies frame the problem in different ways. One approach aims to decompose images into component objects and object features, supporting (among other things) generation of alternative data that permits insertion, deletion, or repositioning of individual objects \citep{burgess2019monet, greff2019multi, engelcke2020genesis, lin2020space}. Another approach aims at a very different form of decomposition---between allocentric scene structure and a variable viewpoint---supporting generation of views of a scene from new vantage points \cite{eslami2018gqn, sitzmann2019srn, mildenhall2020nerf} and, if not supplied as input, estimation of camera pose \cite{Cadena16tro-SLAMfuture}. Although there is work pursuing both of these approaches concurrently in the supervised setting \cite{xu2019mid, nanbo2020mulmon, chen2020object}, very few previous studies have approached the combined challenge in the unsupervised case. In this work, we introduce the Sequence-Integrating Multi-Object Net (\simonet{}), a model which pursues that goal of object-level and viewpoint-level scene decomposition and synthesis without supervision. \simonet\ is designed to handle these challenges without privileged information concerning camera pose, and in dynamic scenes. 

\begin{figure}[t]
    \centering
    \includegraphics[width=0.85\columnwidth]{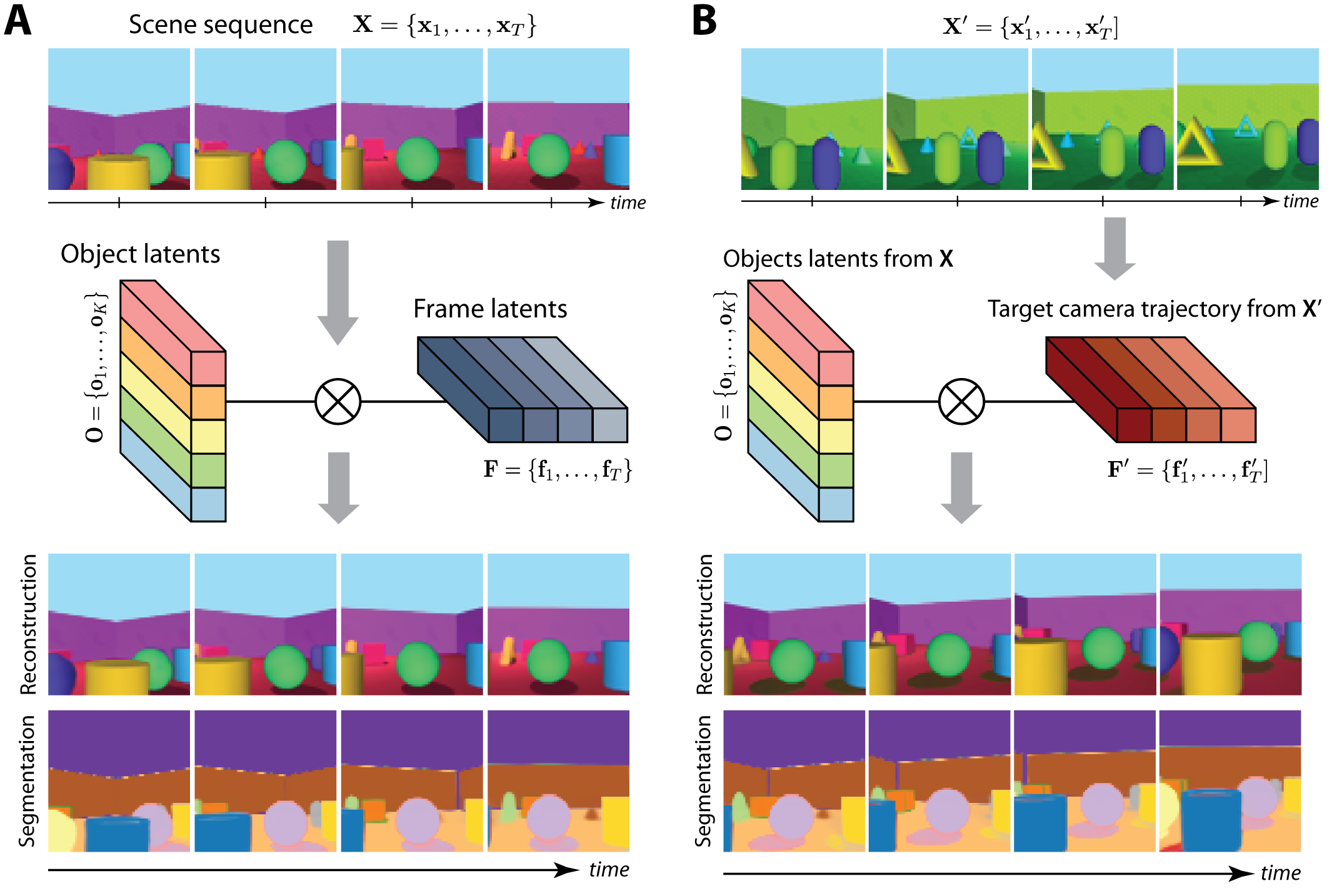}
    \caption{\textbf{Decomposition (A):} \simonet{} factorizes a scene sequence $\mathbf{X}$ into \textit{scene content} (``object latents,'' constant across the sequence) and \textit{view/global content} (``frame latents,'' one per frame) without supervision. Its spatio-temporal attention-based inference naturally allows stable object tracking (e.g. the green sphere is assigned the same segment across frames). \textbf{Recomposition (B):} Object latents of a given sequence $\mathbf{X}$ can be recomposed with the frame latents of a different (i.i.d.) sequence $\mathbf{X}'$ to generate a consistent rendering of the same scene (i.e. objects and their properties, relative arrangements, and segmentation assignments) from entirely different viewpoints. Notice that both camera pose and lighting are transferred, as evidenced by the wall corners in the background and the shadows of the green sphere. % See our \href{\website}{website} for more, animated examples.
    }
    \label{fig:simonet_headline} % \vspace{-10pt}
\end{figure}

Given a video of a scene our model is able to decouple scene structure from viewpoint information (see Figure \ref{fig:simonet_headline}). To do so, it utilizes video-based cues, and a structured latent space which separates time-invariant per-object features from time-varying global features. These features are inferred using a transformer-based network which integrates information jointly across space and time.

Second, our method seeks to summarize objects' dynamics. It learns to disentangle not only static object attributes (and their 2D spatial masks), but also object trajectories, without any prior notion of these objects, from videos alone. The learnt trajectory features are temporally abstract and per object; they are captured independently of the dynamics of camera pose, which being a global property, is captured in the model’s per-frame (time-varying) latents.\footnote{Animated figures are at \href{\website}{\website}.}

Our model thus advances the state of the art in unsupervised, object-centric scene understanding by satisfying the following desiderata: \begin{enumerate*}[label=\textbf{(\arabic*})]
    \item decomposition of multi-object scenes from RGB videos alone;
    \item handling of changing camera pose, and simultaneous inference of scene contents and viewpoint from correlated views (i.e. sequential observations of a moving agent);
    \item learning of structure across diverse scene instances (i.e. procedurally sampled contents);
    \item object representations which summarize static object attributes like color or shape, view-dissociated properties like position or size, as well as time-abstracted trajectory features like direction of motion;
    \item no explicit assumptions of 3D geometry, no explicit dynamics model, no specialized renderer, and few a priori modeling assumptions about the objects being studied; and
    \item simple, scalable modules (for inference and rendering) to enable large-scale use.
\end{enumerate*}

% To summarize, our main contributions are the following:
% \begin{itemize}
%     \item a transformer-based architecture to infer scene contents and views from videos, significantly simplifying the architectures used in prior work.
%     \item a factorized latent space which enforces a separation of static object attributes (including their trajectories) from global/dynamic properties such as camera pose. \al{'Static attributes including trajectories' seems a bit of an oxymoron :) How about something like: a factorized latent space that enforces a separation of individual object properties (both static and dynamic ones) from global properties such as camera pose or global movement}
%     \item experimental results showing competitive view interpolations and video segmentation performance on 3D scenes containing 10 or more objects across three datasets. The datasets are either publicly available or derived from open-source code.
% \end{itemize}

%\vspace{-5pt}
\section{Related Work}
\label{sec:related_work}

Given the multifaceted problem it tackles, \simonet{} connects across several areas of prior work. We describe its nearest neighbors from three scene understanding domains below:

\textbf{Scene decomposition models.} 
\begin{enumerate*}[label=\textbf{(\arabic*})]
    \item Our work builds on a recent surge of interest in unsupervised scene decomposition and understanding, especially using slot structure to capture the objects in a scene \cite{greff2020binding}. One line of work closely related to \simonet{} includes methods like \cite{burgess2019monet, greff2019multi, engelcke2020genesis, lin2020space, locatello2020sa}, which all share \simonet{}'s Gaussian mixture pixel likelihood model. While these prior methods handled only static scenes, more recent work \cite{creswell2021oat, veerapaneni2020entity, zablotskaia2020unsupervised, nanbo2020mulmon} has extended them to videos with promising results. Nevertheless, these approaches have no mechanism or inductive bias to separate view information from scene contents. Moreover, many of them are conditioned on extra inputs like the actions of an agent/camera to simplify inference.

    \item Another family of decomposition models originated with Attend, Infer, Repeat (AIR) \cite{eslami2016air}. AIR's recurrent attention mechanism does split images into components with separate appearance and pose latents each. Later work \cite{lin2020space, kosiorek2018sqair, crawford2019esilot, crawford2019spair, jiang2019scalor} also extended the model to videos. Despite their structured latents, these models do not learn to distill object appearance into a time-invariant representation (as their appearance and pose latents are free to vary as a function of time). They also require separate object discovery and propagation modules to handle appearing/disappearing objects. In contrast, \simonet{} processes a full sequence of images using spatio-temporal attention and produces a single time-invariant latent for each object, hence requiring no explicit transition model or discovery/propagation modules. 
    
    % One particularly interesting descendant of AIR is DDPAE \cite{hsieh2018ddpae}. This model does separate time-varying and time-invariant information like \simonet{}. However, it uses an RNN encoder and spatial transformer-based decoder \citep{jaderberg2015st} (not to be confused with attention-based transformers) like AIR. It has therefore been applied successfully only to simple scenes with no background.
\end{enumerate*}

\textbf{Multi-view scene rendering models.} Models which assume viewpoint information for each image like GQN \cite{eslami2018gqn}, SRNs \cite{sitzmann2019srn}, and NeRF \cite{mildenhall2020nerf} have shown impressive success at learning implicit scene representations and generating novel views from different viewpoints. Recent work \cite{pumarola2020dnerf, park2020nerfies, li2020neuralsceneflow, du2020neuralradianceflow} has further extended these models to videos using deformation fields to model changes in scene geometry over time. In contrast to \simonet{}, these models can achieve photorealistic reconstructions by assuming camera parameters (viewpoint information). To allow a direct comparison, we use a view-supervised version of \simonet{} in Section \ref{sec:results_view_supervised}. There is also recent work \cite{lin2021barf, wang2021nerfmm} that relaxes the known viewpoint constraint, but they still model single scenes at a time, which prevents them from exploiting regularities over multiple scenes. A more recent line of work \cite{kosiorek2021nerfvae, trevithick2020grf, yu2020pixelnerf} explored amortizing inference by mapping from a given set of images to scene latents, but they cannot handle videos yet. Note that all of these models treat the whole scene as a single entity and avoid decomposing it into objects. One exception here is \cite{niemeyer2020giraffe}, which represents objects with separate pose and appearance latents. 
%, and combines these using neural volumetric ray-based rendering. 
However, this model is purely generative and cannot infer object latents from a given scene. Another exception is \cite{chen2020object}, which can in fact infer object representations, but nevertheless depends on view supervision.

\textbf{Simultaneous localization and mapping.} The problem of inferring scene representations in a novel environment by exploring it (rather than assuming given views and viewpoint information) is well studied in robotics and vision \cite{Cadena16tro-SLAMfuture}. Classic SLAM techniques often rely on EM \cite{slam1,slam2} or particle filters \cite{10.5555/1566899.1566949} to infer viewpoint and scene contents jointly. While our problem is slightly simpler (we can leverage shared structure across scene instances; certain elements such as the shape of the room are held constant; and we use offline data rather than active exploration), our approach of using a factorized variational posterior provides a learning-based solution to the same computational problem. Our simplified setting is perhaps justified by our unsupervised take on the problem. On the other hand, we don't assume simplifications which may be common in robotics practice (e.g. known camera properties like field of view; or the use of multiple cameras or depth sensors). Most popular SLAM benchmarks \cite{Burri25012016, Geiger2012CVPR, sturm12iros} are on unstructured 3D scenes and hence it was not straightforward for us to compare directly to classic methods. But an encouraging point of overlap is that object-centric SLAM formulations \cite{xu2019mid} as well as learning-based solutions \cite{9047170, geng2020unsupervised} are active topics of research. Our work could open new avenues in object-centric scene mapping without supervision.

\section{Model}

\simonet{} is a variational auto-encoder \cite{kingma2019introduction} consisting of an inference network (encoder) which infers latent variables from a given input sequence, and a generative process (decoder) which decodes these latents back into pixels. Using the Evidence Lower Bound (ELBO), the model is trained to minimize a pixel reconstruction loss and latent compression KL loss. Crucially, \simonet\ relies on a factorized latent space which enforces a separation of static object attributes from global, dynamic properties such as camera pose. We introduce our latent factorization and generative process in Section \ref{sec:model_latent_structure}. Then in Section \ref{sec:model_inference}, we describe how the latents can be inferred using a transformer-based encoder, significantly simplifying the (recurrent or autoregressive) architectures used in prior work. Finally, we fully specify the training scheme in Section \ref{sec:model_loss}.

% Being a variational auto-encoder, \simonet{} consists of an inference network and a generative model trained simultaneously to reconstruct (a subset of) video pixels while paying a representation cost. Crucially, it relies on a factorized latent space which enforces a separation of static object attributes from global/dynamic properties such as camera pose. We introduce our latent factorization and generative process in Section \ref{sec:model_latent_structure}. Then in Section \ref{sec:model_inference}, we describe how the latents can be inferred using a transformer-based encoder, significantly simplifying the architectures used in prior work. Finally, we specify the training scheme in Section \ref{sec:model_loss}.

%     \item a transformer-based architecture to infer scene contents and views from videos, significantly simplifying the architectures used in prior work.
%     \item a factorized latent space which enforces a separation of static object attributes (including their trajectories) from global/dynamic properties such as camera pose. \al{'Static attributes including trajectories' seems a bit of an oxymoron :) How about something like: a factorized latent space that enforces a separation of individual object properties (both static and dynamic ones) from global properties such as camera pose or global movement}

\subsection{Latent Structure and Generative Process}
\label{sec:model_latent_structure}

% \begin{figure}
%     \centering
%     %\includegraphics{}
%     \caption{Latent structure and generative process of our model.}
%     \label{fig:generative}
% \end{figure}

Our model aims to capture the structure of a scene, observed as a sequence of images from multiple viewpoints (often along a smooth camera trajectory, though this is not a requirement). Like many recently proposed object-centric models we choose to represent the scene as a set of $K$ \emph{object} latent variables $\mathbf{O} \coloneqq \{\mathbf{o}_k\}_{k=1}^K$. These are invariant by construction across all frames in the sequence (i.e. their distribution is constant through time, and expected to summarize information across the whole sequence). We also introduce $T$ \emph{frame} latents $\mathbf{F} \coloneqq \{\view_t\}_{t=1}^T$, one for each frame in the sequence, that capture time-varying information. Note that by choosing this factorization we reduce the number of required latent variables from $K \cdot T$ to $K + T$. The latent prior $p(\mathbf{O}, \mathbf{F})=\prod_k \mathcal{N}(\obj_k \mid \mathbf{0, I})\prod_t  \mathcal{N}(\view_t \mid \mathbf{0, I}) $ is a unit spherical Gaussian, assuming and enforcing independence between object latents, frame latents, and their feature dimensions.

Given the latent variables, we assume all pixels and all frames to be independent. Each pixel is modeled as a Gaussian mixture with $K$ components. The mixture weights for pixel $\mathbf{x}_{t, i}$ capture which component $k$ ``explains'' that pixel ($1 \le i \le HW$). The mixture logits $\hat{\mask}_{k, t, i}$ and RGB (reconstruction) means $\mean_{k, t, i}$ are computed for every component $k$ at a specific time-step $t$ and specific pixel location $\loc_i$ using a decoder $\decoder$:
\begin{align}
    \hat{\mask}_{k,t,i}, \mean_{k,t,i} &= \decoder(\obj_k, \view_t ; \loc_i, t) \\
    p(\mathbf{x}_{t, i} \mid \obj_1, ..., \obj_K, \view_t;t,\loc_i) &= \sum_k \mask_{k, t, i}\mathcal{N}(\mathbf{x}_{t,i} \mid \mean_{k, t, i}; \sigma_x)  \label{eq:pixel_likelihood}
\end{align}
% For designing our slot decoder $\decoder$, we take inspiration from the success of NeRF-like \citep{mildenhall2020nerf} models in representing scenes implicitly.
We decode each pixel independently, "querying" our \emph{pixel-wise} decoder using the sampled latents, coordinates $\loc_i \in [-1, 1]^2$ of the pixel, and time-step $t \in [0, 1)$ being decoded as inputs. The decoder's architecture is an MLP or 1x1 CNN. (See Appendix \ref{app:decoder} for the exact parameterization as well as a diagram of the generative process). By constraining the decoder to work on individual pixels, we can use a subset of pixels as training targets (as opposed to full images; this is elaborated in Section \ref{sec:model_loss}). Once they are decoded, we obtain the mixture weights $\mask_{k,t,i}$ by taking the softmax of the logits across the $K$ components: $\mask_{k,t,i}=\texttt{softmax}_k(\hat{\mask}_{k,t,i})$. Equation \ref{eq:pixel_likelihood} specifies the full pixel likelihood, where $\sigma_x$ is a scalar hyperparameter.

\subsection{Inference}
\label{sec:model_inference}

\begin{figure}
    \centering
    \includegraphics[width=0.9\columnwidth]{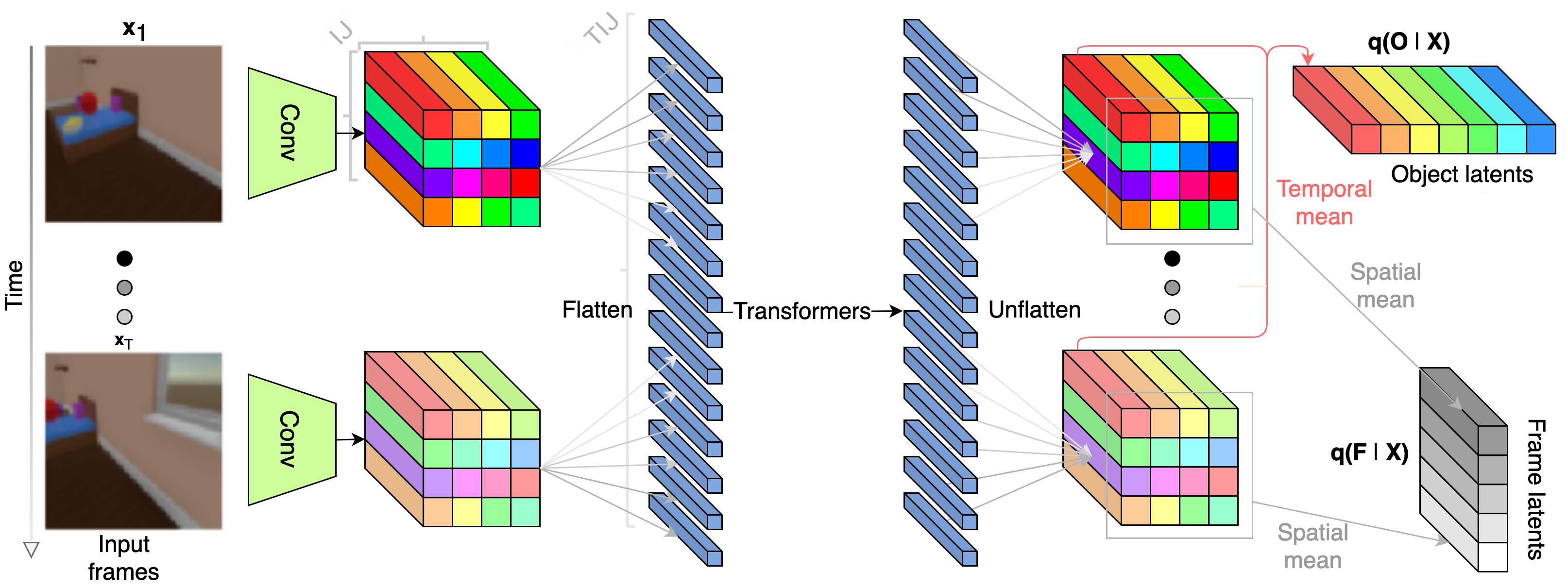}
    \caption{\textbf{Architecture} of the \simonet{} inference network $\encoder$. The transformers integrate information jointly across space and time to infer (the posterior parameters of) the object and frame latents.}
    \label{fig:inference}%\vspace{-10pt}
\end{figure}

Given a sequence of frames $\X \coloneqq \{\x_t\}_{t=1}^T$ we now wish to infer the corresponding object latents $\mathbf{O}$ and frame latents $\mathbf{F}$. The exact posterior distribution $p(\mathbf{O}, \mathbf{F} \mid \X)$ is intractable so we resort to using a Gaussian approximate posterior $q(\mathbf{O}, \mathbf{F} \mid \X)$. The approximate posterior is parameterized as the output of an inference (encoder) network $\encoder(\X)$ which outputs the mean and (diagonal) log scale for all latent variables given the input sequence.

\simonet{}'s inference network is based on the principle that spatio-temporal data can be processed \emph{jointly} across space and time using transformers. Beyond an initial step, we don't need the translation invariance of a CNN, which forces spatial features to interact gradually via a widening receptive field. Nor do we need the temporal invariance of an RNN which forces sequential processing. Instead, we let feature maps interact simultaneously across the cross-product of space and time. See Figure \ref{fig:inference} for an overview of our encoder architecture implementing this.

Concretely, each frame $\x_t$ in the sequence is passed through a CNN which outputs $IJ$ spatial feature maps at each time-step (containing $C$ channels each). $IJ$ can be larger than the number of object latents $K$. (For all results in the paper, we set $I$ and $J$ to 8 each, and $K=16$.) The rest of the inference network consists of two transformers $\mathcal{T}_1$ and $\mathcal{T}_2$. $\mathcal{T}_1$ takes in all $TIJ$ feature maps. Each feature map attends to all others as described. $\mathcal{T}_1$ outputs $TIJ$ transformed feature maps. When $IJ > K$, we apply a spatial pool to reduce the number of slots to $TK$ (see Appendix \ref{app:encoder} for details). These slots serve as the input to $\mathcal{T}_2$, which produces an equal number of output slots. Both transformers use absolute (rather than relative) positional embeddings, but these are 3D to denote the spatio-temporal position of each slot. We denote the output of $\mathcal{T}_2$ as $\mathbf{\hat{e}}_{k, t}$. This intermediate output is aggregated along separate axes (and passed through MLPs) to obtain $T$ frame  and $K$ object posterior parameters respectively. Specifically, $\bm{\lambda}_{\obj_k} = \textrm{mlp}_o(1/T\sum_t \mathbf{\hat{e}}_{k, t})$ while $\bm{\lambda}_{\view_t} = \textrm{mlp}_f(1/K\sum_k \mathbf{\hat{e}}_{k, t})$. Using these posterior parameters we can sample the object latents $\obj_k \sim \mathcal{N}(\bm{\lambda}^{\mu}_{\obj_k}, \exp(\bm{\lambda}^{\sigma}_{\obj_k}) \mathbb{1})$, and the frame latents $\view_t \sim \mathcal{N}(\bm{\lambda}^{\mu}_{\view_t}, \exp(\bm{\lambda}^{\sigma}_{\view_t}) \mathbb{1})$.

% We use a set of frame-wise encoded superpixels $\mathbf{e}_{t, j}$ ($1 \le j \le HW$) as the input to \simonet{}'s inference network $\mathcal{E}_\phi$. The number of superpixels at each time-step can be larger than the number of slots $K$. We set $H$ and $W$ to $2\sqrt{K}$ each.

% \begin{equation}
% \begin{aligned}
%     % \bm{\mu}_k,\, \log \bm{\sigma}_k &= \bm{\lambda}_{\obj_k} \\
%     % \obj_k &\sim \mathcal{N}(\bm{\mu}_k, \bm{\sigma}_k)
%     \obj_k &\sim \mathcal{N}(\bm{\lambda}^{\mu}_{\obj_k}, \exp(\bm{\lambda}^{\sigma}_{\obj_k}) \mathbb{1})
% \end{aligned}
% \end{equation}
% and similarly for the frame latents $\view_t$.

\subsection{Loss and Training}
\label{sec:model_loss}

The model is trained end to end by minimizing the following negative-ELBO derivative:
\begin{equation}
    \begin{aligned}
        \frac{-\alpha}{T_d H_d W_d} \sum_{t=1}^{T_d} \sum_{i=1}^{H_d W_d} \log p(\mathbf{x}_{t, i} \mid \obj_1, ..., \obj_K, \view_t;t,\loc_i) &+ \frac{\beta_o}{K} \sum_k \kldiv{q(\obj_k \mid \X )}{p(\obj_k)} \\
        &+ \frac{\beta_f}{T} \sum_t \kldiv{q(\view_t \mid \X )}{p(\view_t)}
    \end{aligned}
\end{equation}
We normalize the data log-likelihood by the number of decoded pixels $(T_d H_d W_d)$ to allow for decoding fewer than all input pixels ($THW$). This helps scale the size of the decoder (without reducing the learning signal, due to the correlations prevalent between adjacent pixels). Normalizing by $1/T_d H_d W_d$ ensures consistent learning dynamics regardless of the choice of how many pixels are decoded. $\alpha$ is generally set to 1, but available to tweak in case the scale of $\beta_o$ and $\beta_f$ is too small to be numerically stable. Unless explicitly mentioned, we set $\beta_o = \beta_f$. See Appendix \ref{app:our_model} for details.

\section{Comparative Evaluation}
\label{sec:comparative_eval}

To evaluate the model we focus on two tasks: novel view synthesis and video instance segmentation. On the first task (Section \ref{sec:results_view_supervised}), we highlight the benefit of view information when it is provided as ground truth to a simplified version of our model (denoted ``\simonet-VS'' for view supervised), as well as baseline models like GQN \cite{eslami2018gqn} and NeRF-VAE \cite{kosiorek2021nerf}. On the second task (Section \ref{sec:results_segmentation_performance}), we deploy the fully unsupervised version of our model; we showcase not only the possibility of inferring viewpoint from data, but also its benefit to extracting object-level structure in comparison to methods like MONet \cite{burgess2019monet}, Slot Attention \cite{locatello2020sa}, and Sequential IODINE \cite{greff2019multi}.

Our results are based on three procedurally generated video datasets of multi-object scenes. In increasing order of difficulty, they are: \textbf{Objects Room 9} \cite{multiobjectdatasets19}, \textbf{CATER} (moving camera) \cite{Girdhar2020CATER}, and \textbf{Playroom} \cite{abramson2020imitating}. These were chosen to meet a number of criteria: we wanted at least 9-10 objects per scene (there can be fewer in view, or as many as 30 in the case of Playroom). We wanted a moving camera with a randomized initial position (the only exception is CATER, where the camera moves rapidly but is initialized at a fixed position to help localization). We wanted ground-truth object masks to evaluate our results quantitatively. We also wanted richness in terms of lighting, texture, object attributes, and other procedurally sampled elements. Finally, we wanted independently moving objects in one dataset (to evaluate trajectory disentangling and temporal abstraction), and unpredictable camera trajectories in another (the Playroom dataset is sampled using an arbitrary agent policy, so the agent is not always moving). Details on all datasets are in Appendix \ref{app:datasets}.

\subsection{View synthesis (with viewpoint supervision)}
\label{sec:results_view_supervised}

\begin{figure}[t]
    \centering
    \includegraphics[width=\linewidth]{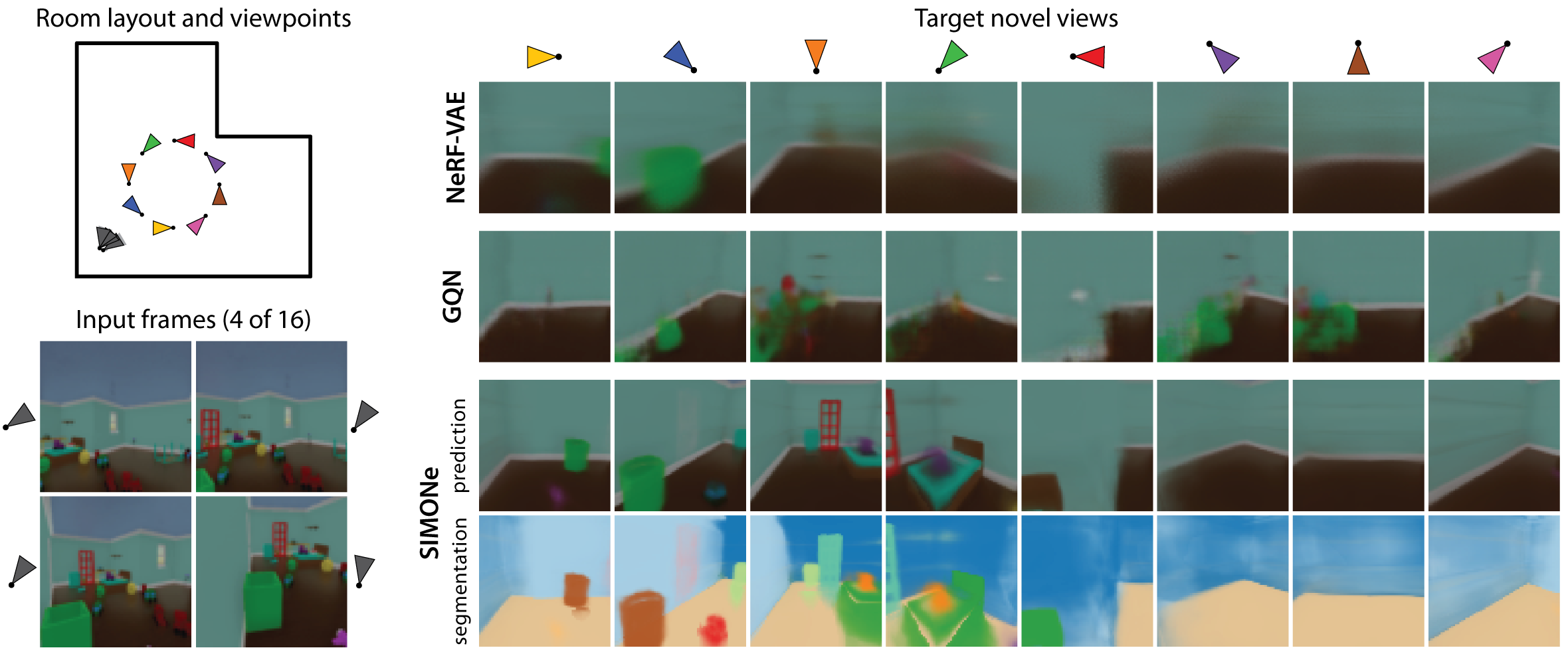}
    \caption{\textbf{Comparison of scene representation and view synthesis capabilities} between \simonet-VS, NeRF-VAE, and GQN. All models partially observe a procedurally generated Playroom from a given sequence of frames (we visualize 4 of the 16 input frames fed to the models). Then, we decode novel views on a circular trajectory around the room, with the yaw linearly spaced in $[-\pi, \pi]$. NeRF-VAE retains very little object structure, while GQN hallucinates content. \simonet-VS can produce fine reconstructions of objects that it observes even partially or at a distance (such as the bed or shelves in the scene). \simonet-VS also segments the scene as a bonus. See Appendix \ref{app:view_supervised_kl_comparison} for similar plots from different scenes/input sequences.}
\label{fig:view_interpolations}%\vspace{-10pt}
\end{figure}

We first motivate view-invariant object representations by considering the case when ground-truth camera pose is provided to our model (a simplified variant we call ``\simonet-VS''). In this scenario, we don't infer any frame latents. Rather, the encoder and decoder are conditioned on the viewpoint directly. This \emph{view-supervised} setting is similar to models like GQN and NeRF which represent the contents of a scene implicitly and can be queried in different directions. 

We compare three such models on view synthesis in the Playroom. The models are provided a set of 16 consecutive frames as context, partially revealing a generated room. Having inferred a scene representation from the input context, the models are then tasked with generating unobserved views of the scene. This extrapolation task is performed without any retraining, and tests the coherence of the models' inferred representations. The task is challenging given the compositional structure of each Playroom scene, as well as the variation across scenes (the color, position, size, and choice of all objects are procedurally sampled per scene; only the L-shaped layout of the room is shared across scenes in the dataset). Because each model is trained on and learns to represent many Playroom instances, NeRF itself is not directly suitable for the task. It needs to be retrained on each scene, whereas we want to infer the specifics of any given room at evaluation time. NeRF-VAE addresses this issue and makes it directly comparable to our model.

To set up the comparison, we first trained \simonet-VS and evaluated its log-likelihood on Playroom sequences. Then, we trained GQN and NeRF-VAE using constrained optimization (GECO \cite{rezende2018taming}) to achieve roughly the same log likelihood per pixel. See Appendix \ref{app:view_supervised_kl_comparison} for a comparison of the models in terms of the reconstruction-compression trade-off.

Qualitatively, the models show vast differences in their perceived structure (see Figure \ref{fig:view_interpolations}). NeRF-VAE blurs out nearly all objects in the scene but understands the geometry of the room and is able to infer wall color. GQN produces more detailed reconstructions, but overfits to particular views and does not interpolate smoothly. \simonet-VS on the other hand finely reproduces the object structure of the room. Even when it observes objects at a distance or up close, it places and sizes them correctly in totally novel views. This makes \simonet-VS a powerful choice over NeRF-VAE and GQN-style models when the priority is to capture scene structure across diverse examples.

% \begin{figure}
%     \centering
%     \includegraphics[width=0.75\columnwidth]{figures/view_supervised_obj_positions.png}
%     \caption{With view supervision, \simonet\ object representations linearly predict the ground-truth allocentric position of each object in the Playroom. This is in contrast to the limitations of existing methods described in Section \ref{sec:intro}. \rk{What's up with the third image? The agent should have more objects in view, given the top-down map.}}
%     \label{fig:view_supervised_obj_positions}
% \end{figure}

\subsection{Instance segmentation (fully unsupervised)}
\label{sec:results_segmentation_performance}

Having shown the benefit of view information to inferring scene structure in the Section \ref{sec:results_view_supervised}, we now turn to the added challenge of inferring viewpoint directly and simultaneously with scene contents (without any supervision).

\begin{figure}
    \centering
    \includegraphics[width=0.82\columnwidth]{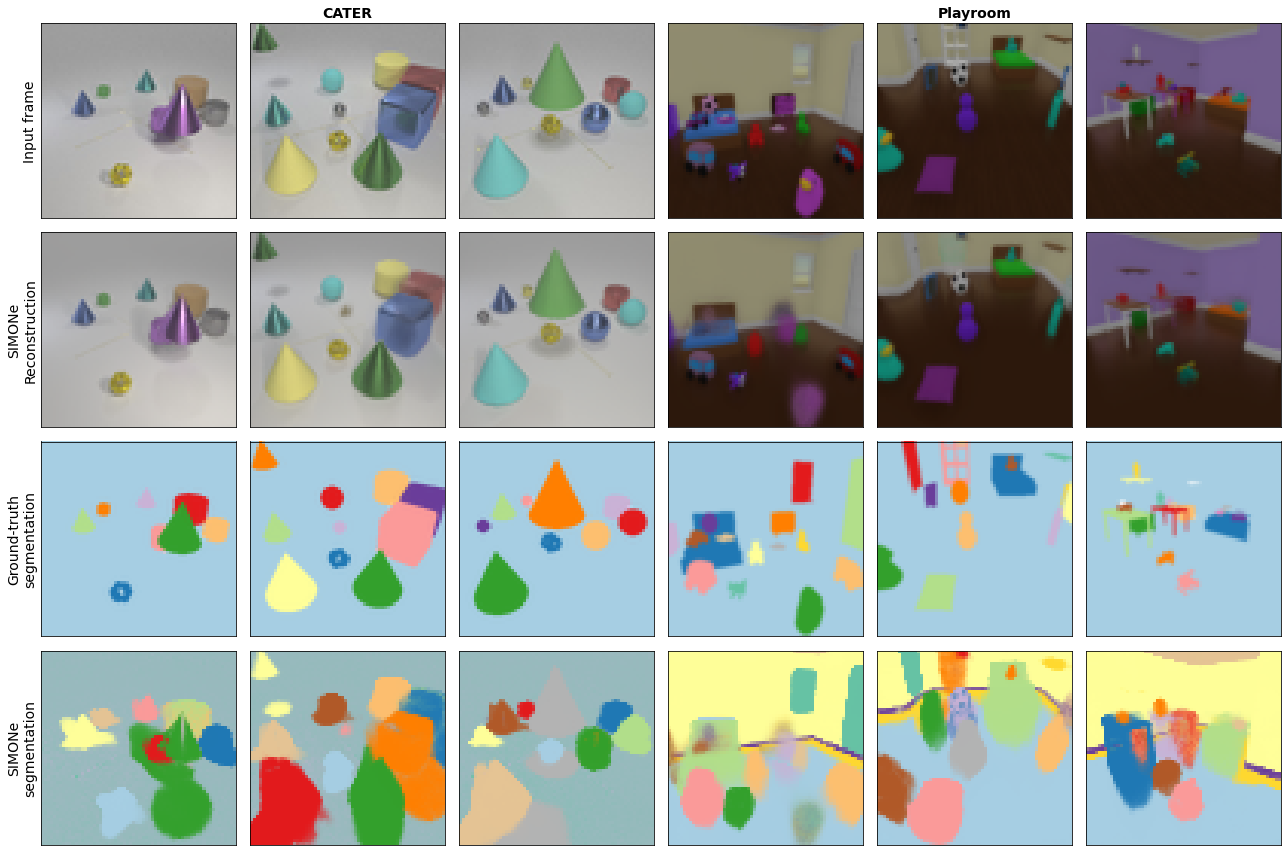}
    \caption{\textbf{Segmentations} and reconstructions produced by \simonet{} on CATER and Playroom. \simonet{} copes well with clutter and different-sized objects. It learns to use object motion as a segmentation signal on CATER, evident from the fact that an object’s shadow is correctly assigned to that object’s segment as it moves. This is true even when there's multiple shadows per object (due to multiple lights in the scene). \simonet{} also overcomes color-based cues to segment two-toned objects such as beds in the Playroom as single objects. See Appendix \ref{app:segmentation_comparison} to compare with baseline models.}
    \label{fig:segmentations_qualitative} %\vspace{-10pt}
\end{figure}

We compare \simonet\ to a range of competitive but viewpoint-unaware scene decomposition approaches. First, we train two static-frame models: MONet and Slot Attention. MONet uses a similar generative process and training loss to our model, achieving segmentation by modeling the scene as a spatial mixture of components, and achieving disentangled representations using a $\beta$-weighted KL information bottleneck. On the other hand, it uses a deterministic, recurrent attention network to infer object masks. Slot Attention is a transformer-based autoencoding model which focuses on segmentation performance rather than representation learning. Finally, we also compare against Sequential IODINE (``S-IODINE''), which applies a refinement network to amortize inference over time, separating objects by processing them in parallel. 
%and breaking symmetries using random initial states
It also uses a $\beta$-weighted KL loss to disentangle object representations. Note that S-IODINE is a simplified version of OP3 \cite{veerapaneni2020entity}, which additionally attempts to model (pairwise) object dynamics using an agent's actions as inputs. \simonet\ and S-IODINE both avoid relying on this privileged information.
Table \ref{tab:segmentation_performance_comparison} contains a quantitative comparison of segmentation performance across these models, while Figure \ref{fig:segmentations_qualitative} shows qualitative results.

\begin{table}[t]
\centering
\footnotesize
\begin{tabular}{llll|llll}
\toprule
 & \multicolumn{3}{c|}{Static ARI-F} & \multicolumn{4}{c}{Video ARI-F} \\
\multicolumn{1}{c}{} & MONet & SA & S-IODINE & MONet & SA & S-IODINE &  SIMONe \\
\midrule
Objects Room 9 & 0.886 & 0.784 & 0.695 & 0.865 & 0.066 & 0.673 & 0.936 \\
& {\scriptsize ($\pm$0.061)} & {\scriptsize ($\pm$0.138)} & {\scriptsize ($\pm$0.007)} & {\scriptsize ($\pm$0.007)} & {\scriptsize ($\pm$0.014)} & {\scriptsize ($\pm$0.0.002)} & {\scriptsize ($\pm$0.010)}\\
CATER & 0.937 & 0.923 & 0.728 & 0.412 & 0.073 & 0.668 & 0.918 \\
& {\scriptsize ($\pm$0.004)} & {\scriptsize ($\pm$0.076)} & {\scriptsize ($\pm$0.032)} & {\scriptsize ($\pm$0.012)} & {\scriptsize ($\pm$0.006)} & {\scriptsize ($\pm$0.033)} & {\scriptsize ($\pm$0.036)}\\
Playroom & 0.647 & 0.653 & 0.439 & 0.442 & 0.059 & 0.356 & 0.800 \\
& {\scriptsize ($\pm$0.012)} & {\scriptsize ($\pm$0.024)} & {\scriptsize ($\pm$0.009)} & {\scriptsize ($\pm$0.010)} & {\scriptsize ($\pm$0.002)} & {\scriptsize ($\pm$0.006)} & {\scriptsize ($\pm$0.043)}\\ \bottomrule
\end{tabular}
\vspace{5pt}
\caption{\textbf{\simonet{} segmentation performance} (in terms of Adjusted Rand Index for foreground objects, ARI-F) compared to state-of-the-art unsupervised baselines: two static-frame models (MONet and Slot Attention, SA) and a video model (S-IODINE). We calculate static and video ARI-F scores separately. For static ARI-F, we evaluate the models per still image. For video ARI-F, we evaluate the models across space and time, taking an object's full trajectory as a single class. The video ARI-F thus penalizes models (especially Slot Attention) which fail to track objects stably. We report the mean and standard deviation of scores across 5 random seeds in each case.
\label{tab:segmentation_performance_comparison}
%\vspace{-10pt}
}
\end{table}

\vspace{-5pt}
\section{Analysis}

We take a closer look at the representations learnt by our model by decoding them in various ways.
% In this section, we assess some characteristics of the representations learnt by our models. 
First, we manipulate latent attributes individually to assess the interpretability of object representations visually in Section \ref{sec:traversals}. Next, we exploit \simonet{}'s latent factorization to render views of a given scene using the camera trajectory of a different input sequence. These cross-over visualizations help identify how the model encodes object dynamics in Section \ref{sec:crosssovers}. Finally, we measure the predictability of ground-truth camera dynamics and object dynamics from the two types of latents in Section \ref{sec:predictions_from_latents}. These analyses use a single, fully unsupervised model per dataset.

\vspace{-5pt}
\subsection{Latent attribute traversals}
\label{sec:traversals}

\begin{figure}[t]
    \centering
    \includegraphics[width=0.75\columnwidth]{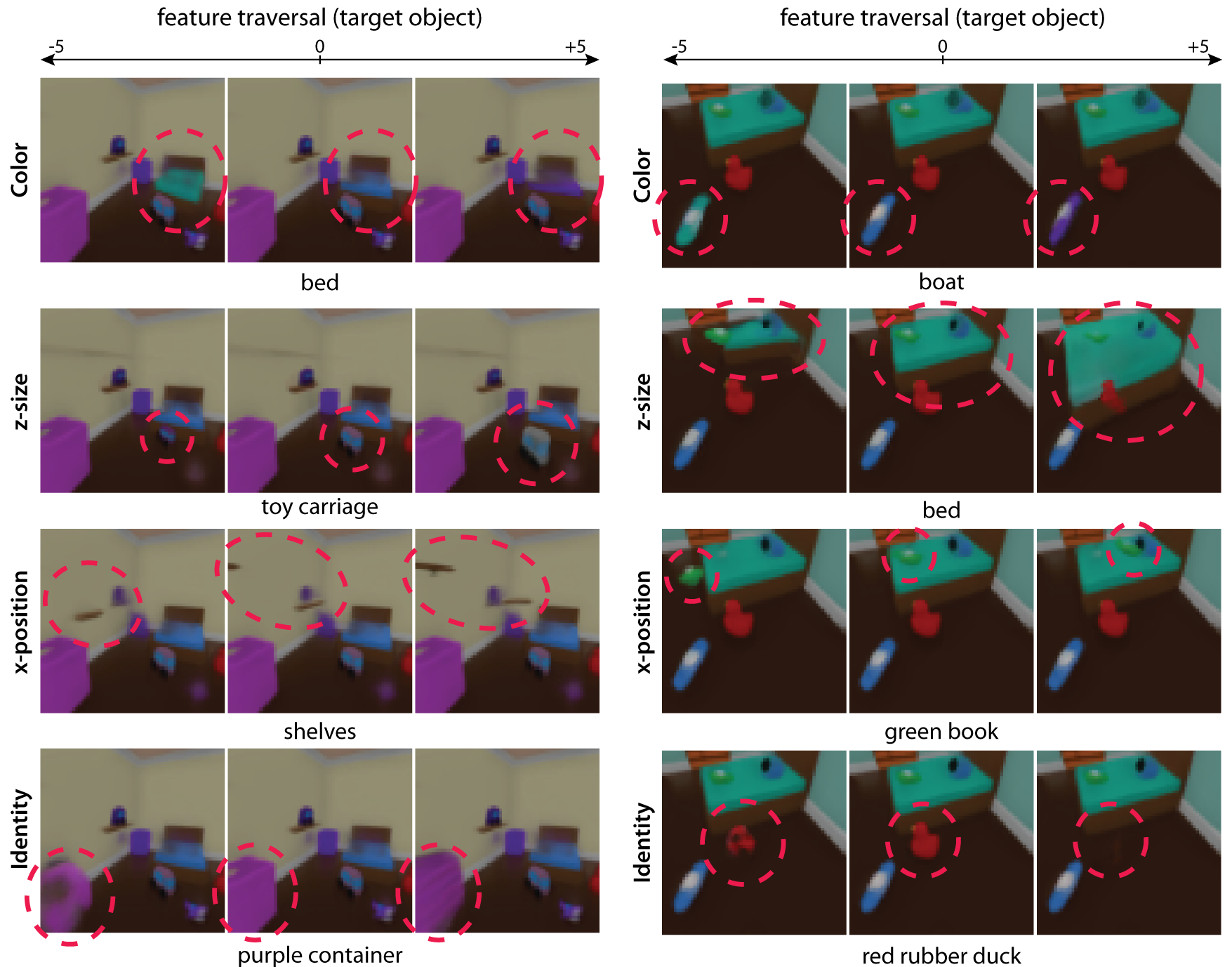}
    \caption{\textbf{Object attributes learnt by \simonet{}.} In each row, we manipulate a particular object latent attribute for an arbitrary target object (circled in red) in two scenes. This reveals the attributes' relationship to interpretable object characteristics like color, size, position and identity.}
    \label{fig:obj_latent_traversals}\vspace{-5pt}
\end{figure}

We visualize the object representations learnt by \simonet{} on Playroom to highlight their disentanglement, across latent attributes and across object slots, in Figure \ref{fig:obj_latent_traversals}. We seed all latents using a given input sequence, then manipulate one object latent attribute at a time by adding fixed offsets.

Note that object position and size are well disentangled in each direction. Aided by the extraction of view-specific information in the frame latents, \simonet{} also learns object features corresponding to identity. The decoder nevertheless obeys the biases in the dataset–for instance, shelves will slide along a wall when their position latent is traversed. The rubber duck does not morph into a chest of drawers because those are always located against a wall. This further suggests a well-structured latent representation, which the decoder can adapt to.

\vspace{-5pt}
\subsection{Object and frame latent cross-overs}
\label{sec:crosssovers}

\begin{figure}
    \centering
    \includegraphics[width=\linewidth]{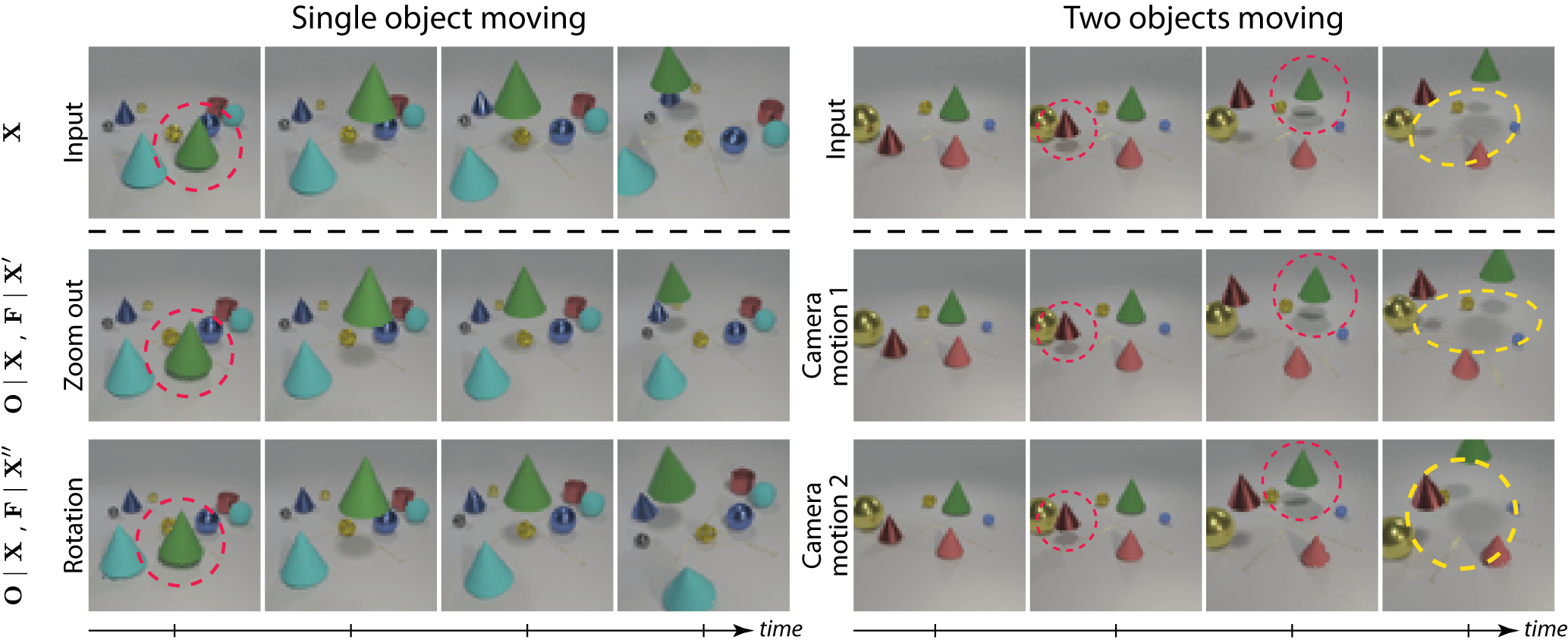}
    \caption{\textbf{Separation of object trajectories from camera trajectories. Left:} When encoding a sequence with consistent (i.i.d.) object dynamics, this information is extracted in the object latents and is unaffected by changing frame latents (see green cone). \textbf{Right:} Movement events are sequenced correctly; object relative positions also remain consistent (see pattern of shadows on the floor circled in yellow). See Appendix~\ref{app:temporal_abstraction} for cross-over plots showing more object trajectories.}
    \label{fig:temporal_abstraction}\vspace{-5pt}
\end{figure}

We expect SIMONe to encode object trajectories and camera trajectories independently of each other. In fact, each object's trajectory should be summarized in its own time-invariant latent code. % This ability–to use an analogy from physics–is like being able to infer the direction of spin of independent electrons (as a static attribute each) by watching a video of them spinning; or extracting wave properties like amplitude and phase while decomposing a superposition (i.e. a mixed signal) observed over time. 
To examine this, we recompose object latents from one sequence with frame latents from other sequences in the CATER dataset. The result, in Figure \ref{fig:temporal_abstraction}, is that we can observe object motion trajectories from multiple camera trajectories. 

Note the consistency of relative object positions (at any time-step) from all camera angles. In the single moving object case, its motion could in fact be interpreted as a time-varying global property of the scene. Despite this challenge, \simonet{} is able to encode the object's motion as desired in its specific time-invariant code. In Section \ref{sec:predictions_from_latents}, we further confirm that object trajectories are summarized in the object latents, which can be queried with time to recover allocentric object positions. 

% Though the object's motion it could be interpreted as a time-varying global scene property in this difficult case where only one object is moving at once

\vspace{-5pt}
\subsection{Camera pose and object trajectory prediction}
\label{sec:predictions_from_latents}

\begin{table}[b!]
\centering
\begin{tabular}{@{}llll@{}}
\toprule
 & Linear($\view_t$) & MLP($\view_t$) & MLP($\obj_1$, ..., $\obj_K$) \\ \midrule
Camera location & $0.832 \pm 0.0$ & $0.949 \pm 0.002$ & $0.044 \pm 0.026$ \\
Camera orientation (Rodrigues) & $0.800 \pm 0.0$ & $0.946 \pm 0.002$ & $0.292 \pm 0.025$ \\ \bottomrule
\end{tabular}
\vspace{10pt}
\caption{\textbf{Decoding camera pose.} We show that ground-truth camera location or orientation is predictable from the corresponding frame latent, but cannot be predicted from all object latents put together. We report the test $R^2$ score across 5 independently trained decoders per input type.}
\label{tab:predicting_camera_pose}%\vspace{-15pt}
\end{table}

\begin{table}[b!]
\centering
\resizebox{\textwidth}{!}{%
\begin{tabular}{@{}lllll@{}}
\toprule
 & MLP($\obj_k$) & MLP($\obj_k, t$) & MLP($\obj_k, \view_t, t$) & MLP($\{\obj_j: j \ne k\}$) \\ \midrule
Trained on all objects & $0.710 \pm 0.006$ & $0.871 \pm 0.006$ & $0.876 \pm 0.003$ & $-0.062 \pm 0.006$\\
Trained on moving objects & $0.724 \pm 0.007$ & $0.894 \pm 0.004$ & $0.898 \pm 0.005$ & $-0.022 \pm 0.025$ \\
\bottomrule
\end{tabular}%
}
\vspace{10pt}
\caption{\textbf{Decoding object trajectories.} We test MLP decoders on predicting allocentric object positions (of moving object in unseen scenes) based on the following inputs: (a) the corresponding object latent, (b) the timestep as well, and (c) the frame latent corresponding to that timestep as well, and (d) remaining object latents from the scene (not pertaining to the object of interest). The decoders were trained on arbitrary objects or a subset containing moving objects only. We report the test $R^2$ score across 5 independently trained decoders per input type.}
\label{tab:sa_predicting_obj_trajectories}%\vspace{-10pt}
\end{table}

We assessed \simonet{}'s frame latents by decoding the true camera position and orientation from them. We trained linear and MLP regressors to predict the camera pose at time $t$ from the corresponding frame latent $\view_t$ on a subset of CATER sequences. We also trained an MLP on the time-invariant object latents $\obj_{1:K}$ for the same task. We evaluated these decoders on held-out data. Table \ref{tab:predicting_camera_pose} shows that frame latents describe the viewpoint almost perfectly.

We also assessed if the object latents contain precise information about allocentric object positions (to be clear, position information is not provided in any form while training \simonet{}). Table~\ref{tab:sa_predicting_obj_trajectories} shows that the correct object latent is predictive of the allocentric position of a dynamic object (when queried along with the timestep). Adding the frame latent does not provide more information, and using the ``wrong'' objects (from the same scene) is completely uninformative. To perform this analysis, we needed to align \simonet's inferred objects with the ground-truth set of objects. We used the Hungarian matching algorithm on the MSE of inferred object masks and ground-truth object masks to perform the alignment. Given \simonet's disentangling of object dynamics, its time-abstracted object representations could prove helpful for a variety of downstream tasks (e.g. ``catch the flying ball!''). 

Taken together, Table~\ref{tab:predicting_camera_pose} and Table~\ref{tab:sa_predicting_obj_trajectories} show the separation of information that is achieved between the object and frame latents, helping assert our two central aims of view-invariant and temporally abstracted object representations.
% \usepackage{booktabs}

%\vspace{-5pt}
\section{Discussion and Future Work}
\label{sec:discussion}

\textbf{Scalability.} The transformer-based inference network in \simonet\ makes it amenable to processing arbitrarily large videos, just as transformer-based language models can process long text. \simonet{} could be trained on windows of consecutive frames sampled from larger videos (aka "chunks"). For inference over a full video, one could add memory slots which carry information over time from one window to the next. Applying \simonet{} on sliding windows of frames also presents the opportunity to amortize inference at any given time-step if the windows are partially overlapping (so the model could observe every given frame as part of two or more sequences). Our use of the standard transformer architecture also makes \simonet\ amenable to performance improvements via alternative implementations.

\textbf{Limitations.} \begin{enumerate*}[label=\textbf{(\arabic*})]
    \item \simonet\ cannot generate novel videos (e.g. sample a natural camera trajectory via consecutive frame latents) in its current version. This could be addressed in a similar fashion to the way GENESIS~\cite{engelcke2020genesis} built on MONet~\cite{burgess2019monet}–it should be possible (e.g. using recurrent networks) to learn conditional priors for objects in a scene and for successive frame latents, which would make \simonet\ fully generative.

    \item We see another possible limitation arising from our strict latent factorization. We have shown that temporally abstracted object features can predict object trajectories when queried by time. This can cover a lot of interesting cases (even multiple object-level "events" over time), but will start to break as object trajectories get more stochastic (i.e. objects transition considerably/chaotically through time). We leave it to future work to explore how temporal abstraction can be combined with explicit per-step dynamics modeling in those cases. For simpler settings, our approach to encoding object trajectories (distilling them across time) is surprisingly effective.
\end{enumerate*}

\section{Conclusion}

We've presented \simonet{}, a latent variable model which separates the time-invariant, object-level properties of a scene video from the time-varying, global properties. Our choice of scalable modules such as transformers for inference, and a pixel-wise decoder, allow the model to extract this information effectively.

\simonet\ can learn the common structure across a variety of procedurally instantiated scenes. This enables it to recognize and generalize to novel scene instances from a handful of correlated views, as we showcased via 360-degree view traversal in the view-supervised setting. More significantly, \simonet{} can learn to infer the two sets of latent variables jointly without supervision. Aided by cross-frame spatio-temporal attention, it achieves state-of-the-art segmentation performance on complex 3D scenes. Our latent factorization (and information bottleneck pressures) further help with learning meaningful object representations. \simonet{} can not only separate static object attributes (like size and position), but it can also separate the dynamics of different objects (as time-invariant localized properties) from global changes in view.

We have discussed how the model can be applied to much longer videos in the future. It also has potential for applications in robotics (e.g. sim-to-real transfer) and reinforcement learning, where view-invariant object information (and summarizing their dynamics) could dramatically improve how agents reason about objects.

\section*{Acknowledgements}

We thank Michael Bloesch, Markus Wulfmeier, Arunkumar Byravan, Claudio Fantacci, and Yusuf Aytar for valuable discussions on the purview of our work. We are also grateful for David Ding's support on the CATER dataset. The authors received no specific funding for this work.

\medskip

{\small
\bibliographystyle{unsrtnat}
\bibliography{egbib}
}
\clearpage
\appendix

\section{Appendix}

\begin{figure}[h]
    \centering
    \includegraphics[width=0.9\columnwidth]{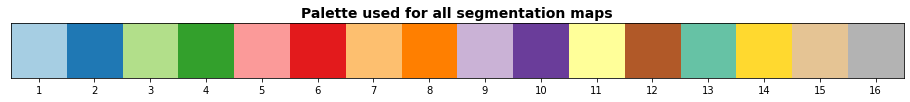}
    \caption{\textbf{Color palette} used for all segmentation masks.}
    \label{fig:color_palette}
\end{figure}

A brief note on notation and convention---we've adopted the following standards for consistency across the paper:
\begin{enumerate}%*}[label=\textbf{(\arabic*})]
    \item Tensor axis order: when indexing into any output of our model (e.g. $\mean_{k,t,i}$), the component aka object dimension (generally size K) precedes the time dimension (size T), which in turn precedes the spatial dimension. We use a single spatial dimension to account for 2D image-centric space.
    \item Allocentric spatial axes: when discussing the position or orientation of an observer or object, we label the axes of 3D space consistently across datasets. The `x` and `z` directions span the ground plane (with z pointing determining an object's depth), while `y` points upward (determining height).
    \item Latent vs latents: We use "latent" to denote a multidimensional hidden variable. Therefore, "latents" denotes a \emph{set} of multidimensional hidden variables (such as $\mathbf{O}$ or $\mathbf{F}$). Any particular dimension of a latent is referred to as a latent attribute or feature.
    \item Segmentation maps: we use a consistent color scheme when plotting segmentation maps (e.g. Figure~\ref{fig:segmentations_qualitative}). See Figure~\ref{fig:color_palette} for our component-wise color palette. The component order is always determined by the output of a given model.
\end{enumerate}%*}

\subsection{Extended Related Work}

\begin{figure}[h]
    \centering
    \includegraphics[width=0.8\columnwidth]{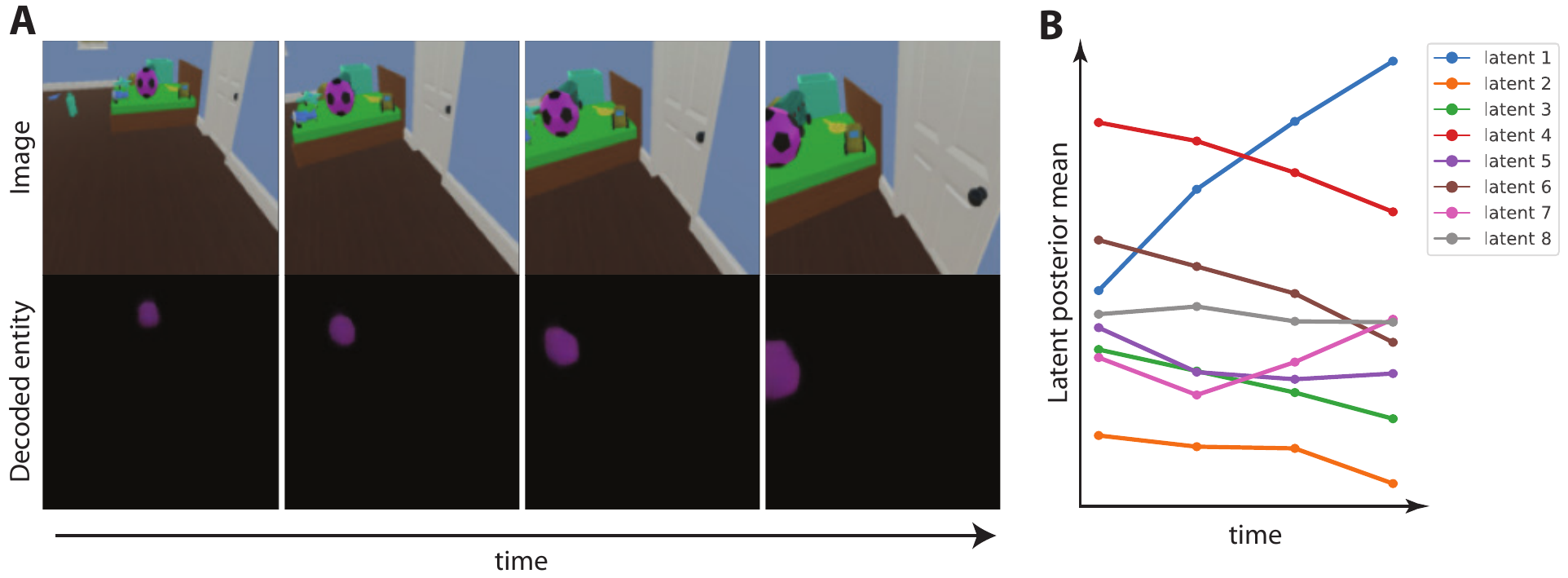}
    \caption{\textbf{Changing view in a 3D scene illustrating an issue with existing models.} \textbf{A.} Consider a static 3D scene observed from a changing viewpoint. We draw attention to the soccer ball lying on the bed. Its size and position appear to change in image space. Row 2 shows the decoded reconstruction of this particular object by a MONet \cite{burgess2019monet} model (fed and trained with ground-truth object masks instead of inferring its own segmentation). \textbf{B.} Given their ignorance of spatial structure with respect to viewpoint, many existing object-centric representation learning methods are bound to conflate changes in viewpoint with object attributes; in this case, MONet encodes the changes directly into the ball's latent representation.}
    \label{fig:changing_view_illustration}
\end{figure}

In Figure \ref{fig:changing_view_illustration}, we demonstrate a limitation of existing scene decomposition approaches which attempt to recognize object-based structure without taking into account a potentially moving observer.

We also draw connections to more prior work (following Section \ref{sec:related_work} in the main paper):

\subparagraph{Video processing and vision.} \begin{enumerate*}[label=\textbf{(\arabic*})]
    \item Human pose estimation has been a common motivation of view-invariant representation learning in vision \cite{parameswaran2006view,liu2018hierarchically,liu2020view,sun2020view}. Some of this work also uses variational unsupervised learning, but the focus is largely on single objects, or makes strong assumptions about the kind of object (e.g. hands and their degrees of freedom \cite{hasson20_handobjectconsist}) being studied.
    
    \item A subset of video understanding and generation models do aim to separate time-invariant and time-varying information like \simonet{}. Some \cite{denton2017uldrv, zhu2020s3vae} require specially designed objectives to achieve this separation; others \cite{denton2018svglp, villegas2017mcnet, yingzhen2018disentangled} structure their generative models to encourage it. While this principle has enabled models to generate quality video predictions, all of them represent the whole scene with a single latent; they cannot decompose it into objects. 
    
    \item Another spate of successes has emerged from the use of transformers \cite{vaswani2017transformer} in vision. They work well at supervised image and video tasks ranging from classification to detection and segmentation \cite{dosovitskiy2021vit, carion2020detr, aang2020maxdeeplab, zheng2020setr}. One such model \cite{caron2021emerging} is capable of foreground-background segmentation without supervision but is trained on still images. For video-based tasks, \cite{bertasius2021timesformer, neimark2021vtn, arnab2021vivit, wang2020vistr} showed the importance of spatio-temporal attention (i.e., integrating information \emph{jointly} over space and time), a principle that also works for \simonet{}'s inference network. However, most prior work relies on supervised learning.
    % The less common, unsupervised use of transformers has focused on image generation \cite{esser2020taming, weissenborn2020scaling}. These models cannot infer the structure of a scene from a video and do not produce segmentations. 
    To our knowledge, \simonet{} is the first model to demonstrate the benefit of spatio-temporal, cross-frame (rather than sequential) attention to decomposing multi-object scenes in a fully unsupervised manner.
\end{enumerate*}

\subsection{Datasets}
\label{app:datasets}  % !!! Has to be A.2 !!!

\subsubsection{Objects Room 9}

Objects Room was a MuJoCo-based dataset originally released \cite{gqn_datasets,multiobjectdatasets19} under the Apache 2.0 license and used for prior work such as GQN and MONet \cite{eslami2018gqn, burgess2019monet}. Our variant, which we denote Objects Room 9, contains more objects per scene (nine rather than three). We use length-16 input sequences with the camera moving on a fixed ring and facing the center of the room. The objects themselves are static.

\subsubsection{CATER}

CATER (a dataset for Compositional Actions and TEmporal Reasoning) was released by \cite{Girdhar2020CATER} under the Apache 2.0 license. We augmented the open-source data generation scripts to further export ground-truth object masks (with lighting disabled, so there's no object shadows). We keep all settings identical to the publicly available version of the dataset containing the moving camera and two moving objects per scene. Like the original dataset, we have three, randomly placed light sources in each scene. This often leads to multiple shadows per object.

\begin{figure}[h]
    \centering
    \includegraphics[width=0.75\columnwidth]{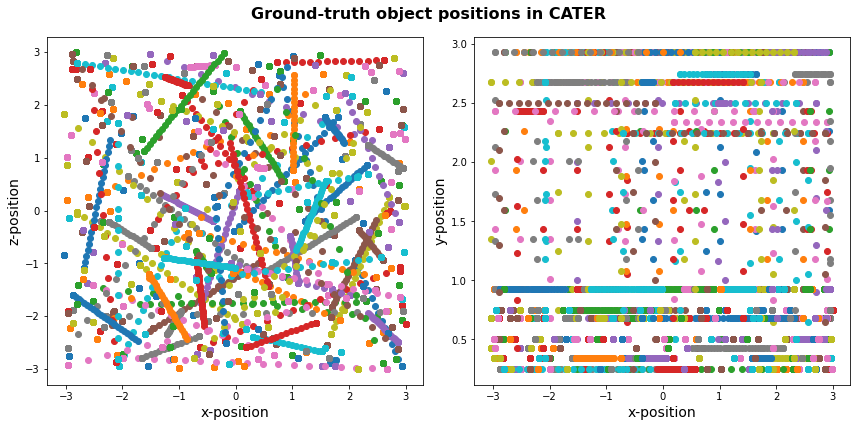}
    \caption{\textbf{Random sample of object positions} from 50 CATER scenes over 32 timesteps.}
    \label{fig:gt_positions_cater}
\end{figure}

See Figure~\ref{fig:gt_positions_cater} for a scatter plot of ground-truth object positions in CATER, highlighting the presence of static and moving objects. 

To train \simonet, we crop the original 320x240 images centrally to a square aspect ratio and then resize them to 64x64. We use length-16 sequences from the beginning of each video. For segmentation figures on CATER (Figures \ref{fig:segmentations_qualitative}, \ref{fig:sa_segmentations_monet}-\ref{fig:sa_segmentations_slotattention}), we add a constant value 0.2 to images of the scene (and reconstructions) to increase their brightness. This is done for visualization only.

\subsubsection{Playroom}

\begin{figure}[h]
    \centering
    \includegraphics[width=0.5\columnwidth]{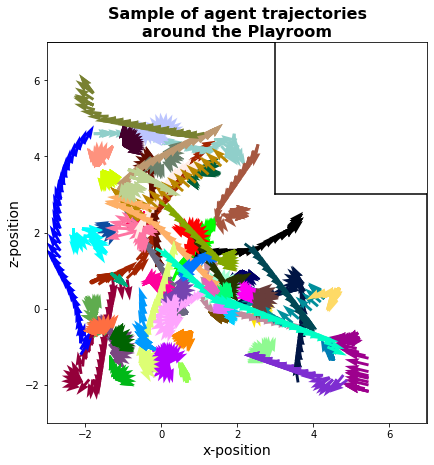}
    \caption{\textbf{Random sample of 64 ground-truth agent trajectories} around the Playroom. Arrow heads denote the agent's orientation, while arrow positions denote location. Note that the trajectories are varied, and the agent rarely observes the full room in a single sequence.}
    \label{fig:gt_positions}
\end{figure}

\begin{figure}[h]
    \centering
    \includegraphics[width=\columnwidth]{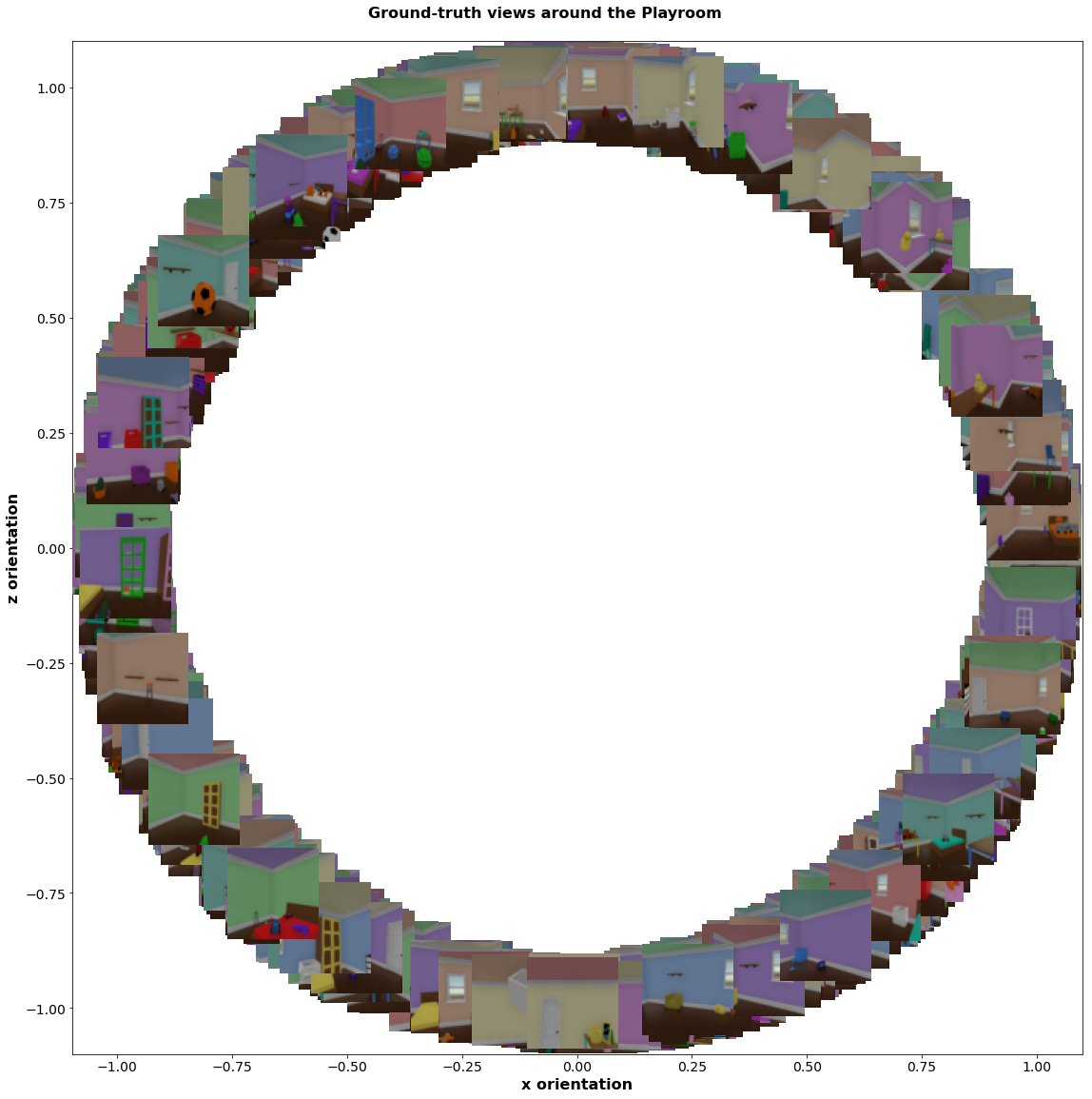}
    \caption{\textbf{A scatter plot of ground-truth views} around the Playroom, when the agent is roughly at the center of the room, with a level pitch (i.e. it's not looking at the floor or ceiling). This reveals the structure and variety in the procedurally generated Playroom from different orientations (yaw).}
    \label{fig:gt_view_interpolation}
\end{figure}

The Playroom is a Unity-based environment for object-centric tasks \cite{hill2020grounded,abramson2020imitating}, originally released as pre-packaged Docker containers with an Apache 2.0 license. We used an arbitrary behaviour policy (trained by demonstrations) to generate video sequences from the environment (one per episode). The L shape of the room remains the same in each instantiation but its appearance and all object/furniture attributes are varied. See Figure \ref{fig:gt_positions} for a sample of the agent's pose in the environment. See also Figure \ref{fig:gt_view_interpolation} for images of Playrooms arranged by the agent's orientation (yaw)---these can be used as a reference for the view synthesis outputs in Figures \ref{fig:view_interpolations}, \ref{fig:sa_view_interpolations_extra_1}, and \ref{fig:sa_view_interpolations_extra_2}.

\clearpage
\subsection{Model and Hyperparameters}
\label{app:our_model}  % !!! Has to be A.3 !!!

See Table~\ref{tab:compute_resources} for compute resources used in training \simonet{} and baseline models.

\subsubsection{Decoder}
\label{app:decoder}

\begin{figure}[h]
    \centering
    \includegraphics[width=0.8\columnwidth]{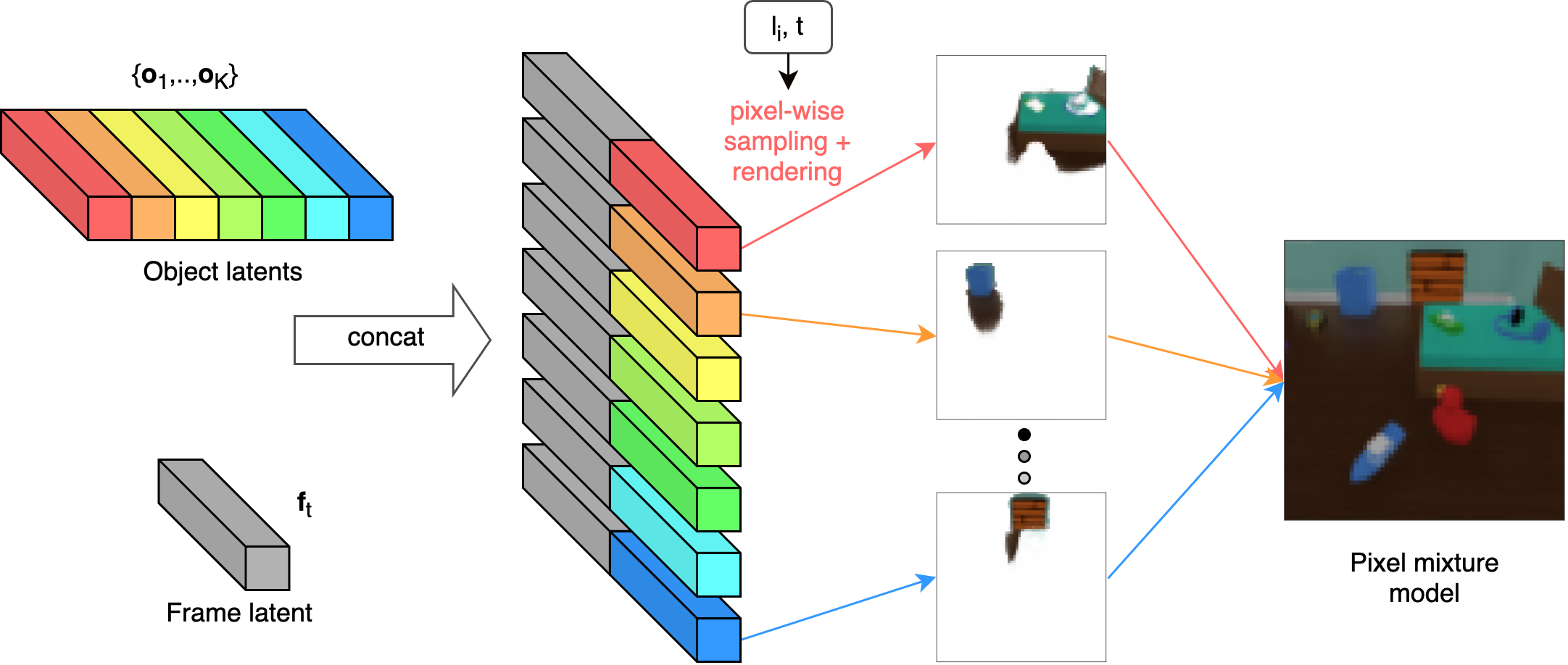}
    \caption{\textbf{\simonet{}'s decoder network} $\decoder$ architecture.}
    \label{fig:generative}
\end{figure}

Our pixel-level decoder $\decoder(\obj_k, \view_t ; \loc_i, t)$ is implemented as a 1x1 CNN (effectively a pixel-wise MLP), which receives concatenated inputs $[\obj_k, \view_t, \loc_i, t]$. Note this means the decoder can be parallelized (i.e. batch-applied) across all batch elements, all K objects, and all T frames.  For CATER and Playroom, the CNN has 6 hidden layers with 512 output channels each. For Objects Room 9, we use 4 hidden layers with 128 channels each. An additional output layer produces 4 channels containing the RGB pixel (reconstruction) means $\mean_{k, t, i}$ and pre-norm mixture logits. The final mixture logits $\hat{\mask}_{k, t, i}$ are derived by applying layer norm to the pre-norm mixture logits on the $\{k, t, i\}$ axes simultaneously.

We've used $\decoder(\obj_k, \view_t ; \loc_i, t)$ as shorthand for $\decoder(\obj_{k,t,i}, \view_{k,t,i}; \loc_i, t)$. Instead of prior decoding approaches \cite{watters2019spatial} which sampled a single latent variable per object and broadcast it spatially to form the input to the decoder, we take i.i.d samples $\obj_{k, t, i} \sim q(\obj_k | \X)$ across all time-steps $1 \le t \le T$ and across all pixels $1 \le i \le HW = 64 \cdot 64$. Likewise, we take i.i.d samples $\view_{k, t, i} \sim q(\view_t | \X)$ across all object slots $1 \le k \le K$ and across all pixels $1 \le i \le HW = 64 \cdot 64$. This "multisampling" approach, while not crucial, does improve early training performance. We hypothesize that this is due to a reduction in the gradient bias resulting from (the alternative approach of) copying single samples $\obj_k$ and $\view_t$ across all space and time in a sequence. % Taking multiple independent samples acts as a regularizer.

Hence, to be explicit, the full pixel-wise decoder can be written as follows:
$$\hat{\mask}_{k,t,i}, \mean_{k,t,i} = \decoder(\obj_{k,t,i}, \view_{k,t,i} ; \loc_i, t)$$

% We explored different ways to use the latents to enable better pixel reconstruction. In particular, we first started from an architecture similar to the spatial broadcast decoder \cite{watters2019spatial}, where the sample $\obj_k$ (along with $\view_t$ in our case) is first copied and tiled spatially across all pixels and time points before using a convolutional decoder.

% We found that alternatively, one can take independent samples in order to seed the first stage of the decoder, before applying a 1x1 convolutional decoder stack. We additionally independently sample across time when appropriate. 
% In summary, we use samples $\obj_{k, i, t} \sim q(o_k | \X)$ (independently for all pixels $I=64 \cdot 64$ and time frames $T$) and $\view_{k, t, i} \sim q(f_t | \X)$ (independently for all slots $K$ and all pixels $I=64 \cdot 64$).
% We found that this improved performance, especially early in training. We hypothesize that this is due to a reduction in the gradient bias involved with copying the same samples $\obj_k$ and $\view_t$ across all space/time in a sequence. Taking multiple independent samples acts as a regularizer.

\subsubsection{Encoder}
\label{app:encoder}

See Figure~\ref{fig:inference} in the main text for an architecture diagram.

\subparagraph{CNN.} The encoder $\encoder(\X)$ contains an initial CNN which outputs $IJ = 8 \cdot 8$ spatial feature maps for each frame. The CNN layers have stride 2, kernel size 4, and output 128 channels each---hence the number of layers is determined by the size of the input (e.g. three layers when the input size $HW = 64 \cdot 64$). We use a ReLU activation after each layer.

\subparagraph{Transformers.} The CNN is followed by transformers $\mathcal{T}_1$ and $\mathcal{T}_2$, which have identical hyperparameters. For CATER and Playroom, the transformers have 4 layers, 5 heads, value size 64, and MLPs with a single hidden layer (size 1024). For Objects Room 9, we use even simpler transformers, with 3 heads and 256 MLP hidden units each (other settings are kept the same). Note that the transformer embedding size is determined as the product of the number of heads and the value size---it is not constrained by the dimensionality of the transformer inputs. We don't use dropout but do use layer norm (on the input) in each transformer residual block.

\subparagraph{Spatial pool (if necessary).} In between $\mathcal{T}_1$ and $\mathcal{T}_2$, we need to reduce the number of slots from $TIJ$ to $TK$ (if $IJ > K$). We use a strided \emph{sum} across the spatial dimensions (sized $I$ and $J$) to do so, with the kernel size and stride each set to $[I/\sqrt{K}, J/\sqrt{K}] = [2, 2]$. Finally, we scale the pooled values by $\sqrt{K/IJ} = \sqrt{16/64} = 1/2$ for our experiments.

\subparagraph{MLPs.} The outputs $\mathbf{\hat{e}}_{k, t}$ of $\mathcal{T}_2$ are aggregated separately along the spatial and temporal axes. These aggregates are passed through $\textrm{mlp}_f$ and $\textrm{mlp}_o$ to yield the frame and object posterior parameters respectively. Both MLPs have a single hidden layer with 1024 units.

\begin{table}[t]
\centering
{\renewcommand{\arraystretch}{1.2}
\begin{tabular}{|p{4cm}|P{2.5cm}|P{2.5cm}|P{2.8cm}|}
\hline
 & \textbf{Objects Room 9} & \multicolumn{1}{c|}{\textbf{CATER}} & \multicolumn{1}{c|}{\textbf{Playroom}} \\ \hline
Shape of input, {[}T, H, W{]} & {[}16, 64, 64{]} & {[}16, 64, 64{]} & {[}32, 64, 64{]} \\ \hline
Shape of decoded image, {[}$T_d$, $H_d$, $W_d${]} & {[}4, 32, 32{]} & {[}8, 32, 32{]} & {[}4, 64, 64{]} \\ \hline
Number of objects, K & \multicolumn{3}{c|}{16} \\ \hline
Frame/object latent dimensionality & \multicolumn{3}{c|}{32} \\ \hline 
Object latents KL loss weight, $\beta_o$ & 1e-8 & 1e-5 & \begin{tabular}[c]{@{}c@{}}Annealed: \\ 5e-6 $\rightarrow$ 5e-7\\ (exponential\\ window: 50k steps)\end{tabular} \\ \hline
Frame latents KL loss weight, $\beta_f$ & 1e-8 & 1e-4 & 1e-7 \\ \hline
Reconstruction (NLL) loss scale, $\alpha$ & \multicolumn{3}{c|}{0.2} \\ \hline
Pixel likelihood scale, $\sigma_x$ & \multicolumn{3}{c|}{0.08} \\ \hline
Number of training iterations & 4e5 & 2.5e6 & 2.5e6 \\ \hline
Learning rate & \multicolumn{3}{c|}{20e-5} \\ \hline
Optimizer & \multicolumn{3}{c|}{Adam} \\ \hline
\end{tabular}
}
\vspace{5pt}
\caption{\textbf{Summary of hyperparameters} for SIMONe across all datasets.}
\label{tab:simone_hyperparams}\vspace{-15pt}
\end{table}

\subsubsection{Training details}
While training, we decode fewer frames than the number fed and encoded by the model as motivated in Section \ref{sec:model_loss}. We simply take $T_d$ random frame indices from $\{1, ..., T\}$ without replacement. However, when we subsample pixels to decode, we use a strided slice of the input (e.g. stride [2, 2] when $H_d = H/2$  and $W_d = W/2$). Note that for any evaluation or visualization, we decode the full length and size $[T, H, W]$ of the input sequence.

\subsection{Baseline Models}
\label{app:baseline_models}  % !!! Has to be A.4 !!!

See Table~\ref{tab:compute_resources} for compute resources used in training baseline models.

\subsubsection{Slot Attention}
\label{app:slot_attention}

For all three datasets, we use slot size (i.e., dimensionality) 32 and train for 500,000 steps with a batch size of 128. We use linear learning rate warmup over the first 10,000 steps. Slot Attention seemed very sensitive to the choice of number of slots, with the model prone to oversegmenting rather than leaving slots empty. So we swept over the following settings for all datasets: number of slots (7 vs 10), encoder/decoder architecture (see Table~\ref{tab:sa_architecture}), and learning rate ($\expnumber{4}{-3}, \expnumber{4}{-4}, \expnumber{4}{-5}$). We ran 5 random seeds for each hyperparameter setting. The mean and standard deviations reported in the Table~\ref{tab:segmentation_performance_comparison} are calculated over the random seeds for the best performing hyperparameter setting (in terms of ARI-F).

\subsubsection{MONet}
\label{app:monet}

For all three datasets, we use 10 object slots, 64-dimensional latents, and train for 5,000,000 steps with an effective batch size of 32. We swept over the following: scale of the KL penalty $\beta$ (0.5 or annealed from 0.01 to 30 with an exponential window of 200,000 steps), pixel likelihood scale for the foreground slots (0.08 or 0.09), and the size of the pixel broadcast decoder (64 channels/5 hidden layers or 512 channels/6 hidden layers). The decoder settings are the same we used for the Slot Attention sweeps, except that we use 1x1 convolutions with stride 1 for MONet. Note that the background pixel likelihood scale is fixed at 0.07. The encoder is identical to the original implementation. 

\subsubsection{S-IODINE}
\label{app:iodine}

For all three datasets, we use latent size of 64 and train for 500,000 steps with a batch size of 128. We use a fixed learning rate of 3e-4 using the Adam optimizer. We use 7 slots for all models. We swept over the following: encoder/decoder architecture (see Table~\ref{tab:iodine_architecture}), scale of KL term (0.5 or 1.0) and output likelihood standard deviation (0.08 and 0.09).

\subsubsection{GQN}
For the view synthesis comparison, we trained GQN on Playroom data. The architecture is the same as in \cite{kosiorek2021nerfvae}. We use an autoregressive decoder with 5 steps, 16-dimensional latents, and 256 hidden units in each layer. We use a Nouveau ResNet encoder and a 2D convolutional LSTM as the recurrent core. We set the likelihood scale to 0.08 to ensure the output distribution is parameterized identically to \simonet-VS. To train the model, we used GECO to target a minimum reconstruction log likelihood (4.3, 4.5 or 4.7 nats per pixel). We used 16 frames as the context for each scene; these were randomly sampled from a 32-frame sequence.

\subsubsection{NeRF-VAE}

The NeRF-VAE model for the view synthesis comparison is the same as in \cite{kosiorek2021nerfvae}. Like GQN, we trained NeRF-VAE with GECO, setting thresholds of 3.8, 4.0 or 4.2 nats per pixel for the reconstruction log likelihood (values higher than 3.8 were not attained). We used 16 context frames from the sequence, 0.08 likelihood scale, and 512 latent dimensions.

\begin{table}[t]
\centering
{\renewcommand{\arraystretch}{1.2}
\footnotesize
\begin{tabular}{ll|c|c|}
\cline{3-4}
 &  & \textbf{Small} & \textbf{Large} \\ \hline
\multicolumn{1}{|l|}{\multirow{3}{*}{Encoder}} & CNN & \begin{tabular}[c]{@{}l@{}}Conv2D(c=64, k=5, s=1)\\ Conv2D(c=64, k=5, s=1)\\ Conv2D(c=64, k=5, s=1)\\ Conv2D(c=64, k=5, s=1)\end{tabular} & \begin{tabular}[c]{@{}l@{}}Conv2D(c=512, k=5, s=1)\\ Conv2D(c=512, k=5, s=1)\\ Conv2D(c=512, k=5, s=1)\\ Conv2D(c=512, k=5, s=1)\\ Conv2D(c=512, k=5, s=1)\\ Conv2D(c=512, k=5, s=1)\end{tabular} \\ \cline{2-4} 
\multicolumn{1}{|l|}{} & Position MLP & MLP(64) & MLP(512) \\ \cline{2-4} 
\multicolumn{1}{|l|}{} & Output MLP & MLP({[}64, 64{]}) & MLP({[}512, 512{]}) \\ \hline
\multicolumn{1}{|l|}{} & Slot Attention & \begin{tabular}[c]{@{}c@{}}num iterations=3, slot size=32\\ GRU(32)\\ MLP(128)\end{tabular} & \begin{tabular}[c]{@{}c@{}}num iterations=3, slot size=32\\ GRU(32)\\ MLP(512)\end{tabular} \\ \hline
\multicolumn{1}{|l|}{\multirow{2}{*}{Decoder}} & CNN & \begin{tabular}[c]{@{}l@{}}Conv2D$^T$(c=64, k=5, s=2)\\ Conv2D$^T$(c=64, k=5, s=2)\\ Conv2D$^T$(c=64, k=5, s=2)\\ Conv2D$^T$(c=64, k=5, s=1)\\ Conv2D$^T$(c=64, k=5, s=1)\\ Conv2D(c=4, k=3, s=1)\end{tabular} & \begin{tabular}[c]{@{}l@{}}Conv2D$^T$(c=512, k=1, s=1)\\ Conv2D$^T$(c=512, k=1, s=1)\\ Conv2D$^T$(c=512, k=1, s=1)\\ Conv2D$^T$(c=512, k=1, s=1)\\ Conv2D$^T$(c=512, k=1, s=1)\\ Conv2D$^T$(c=512, k=1, s=1)\\ Conv2D(c=4, k=3, s=1)\end{tabular} \\ \cline{2-4} 
\multicolumn{1}{|l|}{} & Position MLP & MLP({[}64, 32{]}) & MLP({[}128, 32{]}) \\ \hline
\end{tabular}
}
\vspace{5pt}
\caption{\textbf{Slot Attention baseline architectures.} c: number of channels, k: kernel size, s: stride, Position MLP: MLP applied to positional encoding. MLP([m, n]) denotes an MLP with layers of size m and n.}
\label{tab:sa_architecture}\vspace{-15pt}
\end{table}

\begin{table}[h]
\centering
{\renewcommand{\arraystretch}{1.2}
\footnotesize
\begin{tabular}{ll|c|c|}
\cline{3-4}
 &  & \textbf{Small} & \textbf{Large} \\ \hline
\multicolumn{1}{|l|}{\multirow{3}{*}{Encoder}} & CNN & \begin{tabular}[c]{@{}l@{}}Conv2D(c=32, k=3, s=2)\\ Conv2D(c=32, k=3, s=2)\\ Conv2D(c=32, k=3, s=2)\\ Conv2D(c=64, k=3, s=2)\end{tabular} & \begin{tabular}[c]{@{}l@{}}Conv2D(c=64, k=5, s=1)\\ Conv2D(c=64, k=5, s=1)\\ Conv2D(c=64, k=5, s=1)\\ Conv2D(c=64, k=5, s=1)\\ Conv2D(c=64, k=5, s=1)\end{tabular} \\ \hline % \begin{tabular}[c]{@{}l@{}}Conv2D(c=32, k=3, s=2)\\ Conv2D(c=32, k=3, s=2)\\ Conv2D(c=64, k=3, s=2)\\ Conv2D(c=64, k=3, s=1)\end{tabular} \\ 
% \cline{2-4} 
\multicolumn{1}{|l|}{} & Output MLP & \multicolumn{2}{c|}{MLP({[}256, 256{]})} \\ \hline
\multicolumn{1}{|l|}{} & Refinement Network & \multicolumn{2}{c|}{\begin{tabular}[c]{@{}l@{}} LSTM(128)\\ Linear(128)\end{tabular}} \\ \cline{1-4} %
\multicolumn{1}{|l|}{\multirow{2}{*}{Decoder}} & CNN & \begin{tabular}[c]{@{}l@{}}Conv2D$^T$(c=32, k=5, s=2)\\ Conv2D$^T$(c=32, k=5, s=2)\\ Conv2D$^T$(c=32, k=5, s=2)\\ Conv2D$^T$(c=32, k=5, s=1) \\ Conv2D(c=4, k=3, s=1)\end{tabular} & \begin{tabular}[c]{@{}l@{}}Conv2D$^T$(c=512, k=1, s=1)\\ Conv2D$^T$(c=512, k=1, s=1)\\ Conv2D$^T$(c=512, k=1, s=1)\\ Conv2D$^T$(c=512, k=1, s=1)\\ Conv2D$^T$(c=512, k=1, s=1)\\ Conv2D(c=4, k=3, s=1)\end{tabular} \\ \cline{2-4} 
\hline
\end{tabular}
}
\vspace{5pt}
\caption{\textbf{S-IODINE baseline architectures.} c: number of channels, k: kernel size, s: stride, MLP([m, n]) denotes an MLP with layers of size m and n.}
\label{tab:iodine_architecture}
\end{table}

\begin{table}[h]
\resizebox{\textwidth}{!}{%
\begin{tabular}{llll}
 & \textbf{Playroom} & \textbf{CATER} & \textbf{Objects Room} \\ \hline
SIMONe & 64/128 TPUv1 $\times$ 9 sweeps $\times$ 5 reps & 64 TPUv2 $\times$ 4 sweeps $\times$ 5 reps & 64 TPUv2 $\times$ 4 sweeps $\times$ 5 reps \\
MONet & 32 TPUv1 $\times$ 8 sweeps $\times$ 5 reps & 32 TPUv1 $\times$ 8 sweeps $\times$ 5 reps & 32 TPUv1 $\times$ 8 sweeps $\times$ 5 reps \\
S-IODINE & 32 TPUv1 $\times$ 16 sweeps $\times$ 5 reps & 32 TPUv1 $\times$ 16 sweeps $\times$ 5 reps & 32 TPUv1 $\times$ 16 sweeps $\times$ 5 reps \\
Slot Attention & 8 TPUv2 $\times$ 24 sweeps $\times$ 5 reps & 8 TPUv2 $\times$ 24 sweeps $\times$ 5 reps & 8 TPUv2 $\times$ 24 sweeps $\times$ 5 reps
\end{tabular}
}
\vspace{5pt}
\caption{\textbf{Compute resources} used to train \simonet{} and baseline models. TPU: Tensor Processing Unit. TPUv1 and TPUv2 have 8GiB and 16GiB memory respectively. Each TPU unit refers to 1 core. Each "sweep" refers to a unique hyperparameter combination, while "reps" refer to independent random seeds.}
\label{tab:compute_resources}\vspace{-10pt}
\end{table}

\subsection{Extended Results}
\label{app:results}

\subsubsection{View synthesis}
\label{app:view_supervised_kl_comparison}

\begin{figure}[h]
    \includegraphics[width=\linewidth]{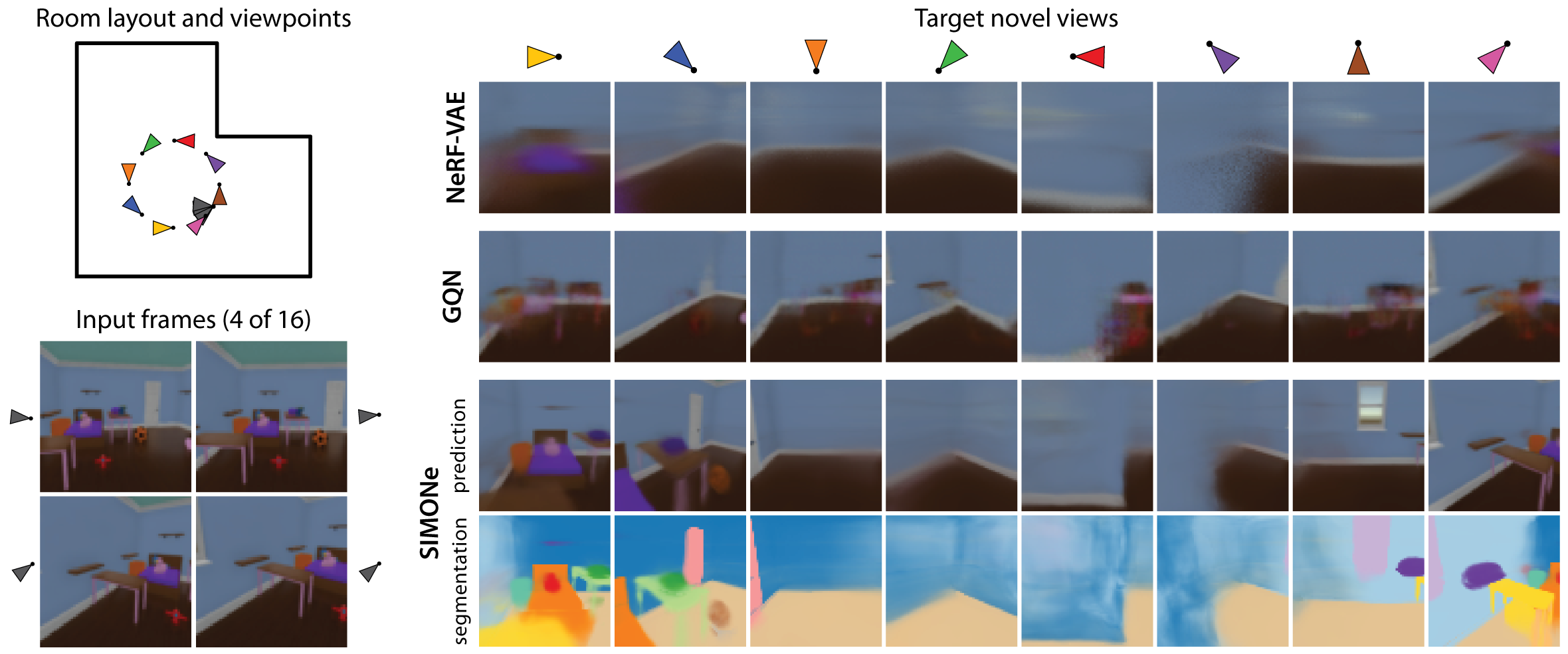}\\
    \vspace{5pt}
    \includegraphics[width=\linewidth]{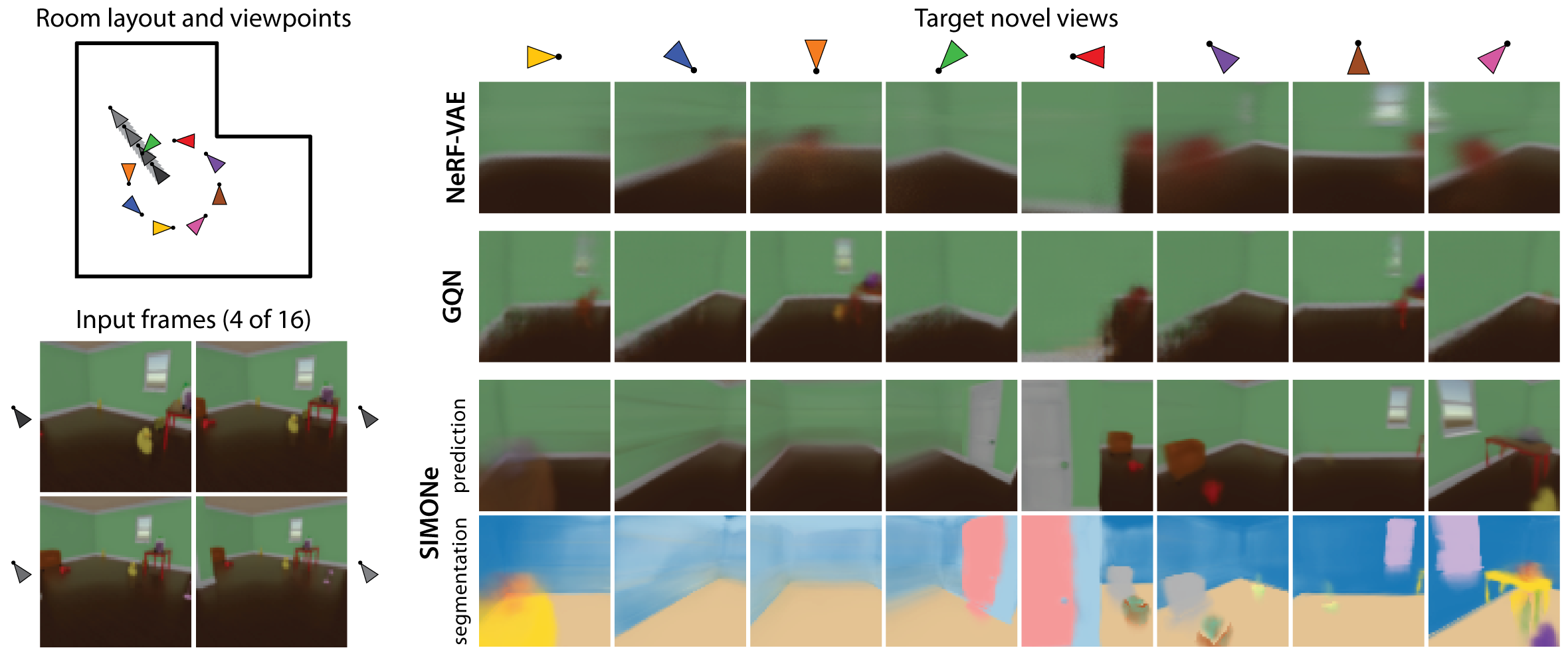}\\
    \vspace{5pt}
    \includegraphics[width=\linewidth]{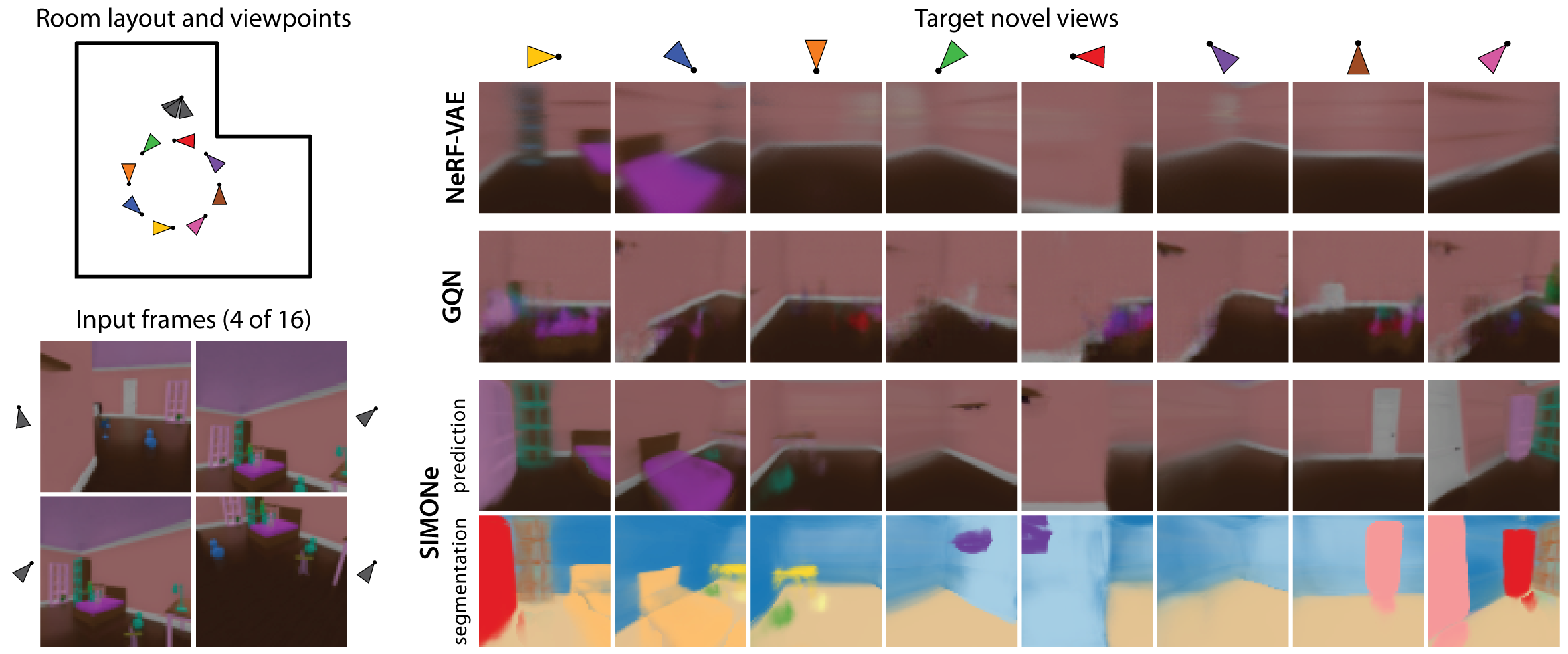}
    \caption{\textbf{Extended comparisons of scene representation and view synthesis capabilities} between \simonet-VS, NeRF-VAE, and GQN. In each figure, a sub-sample of the input context sequence is shown at the lower left. At the upper left, a map of the room shows the camera pose corresponding to the visualized input images in dark gray, as well as the remaining input frames (observed by the models) in light gray. Columns on the right correspond to the colored (novel) viewpoints on the map. Refer to Figure~\ref{fig:view_interpolations} for more details.}
\label{fig:sa_view_interpolations_extra_1}
\end{figure}

\begin{figure}[t]
    \centering
    \includegraphics[width=\linewidth]{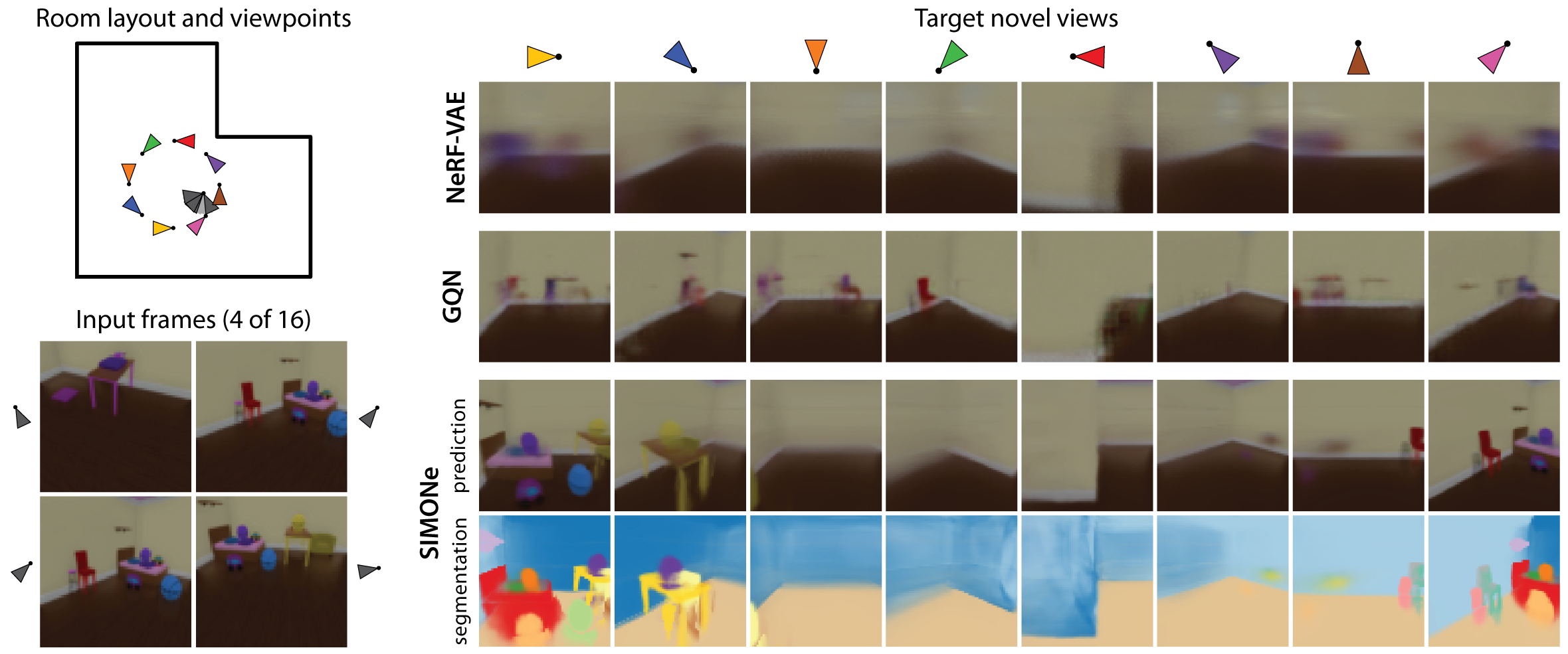}\\
    \vspace{5pt}
    \includegraphics[width=\linewidth]{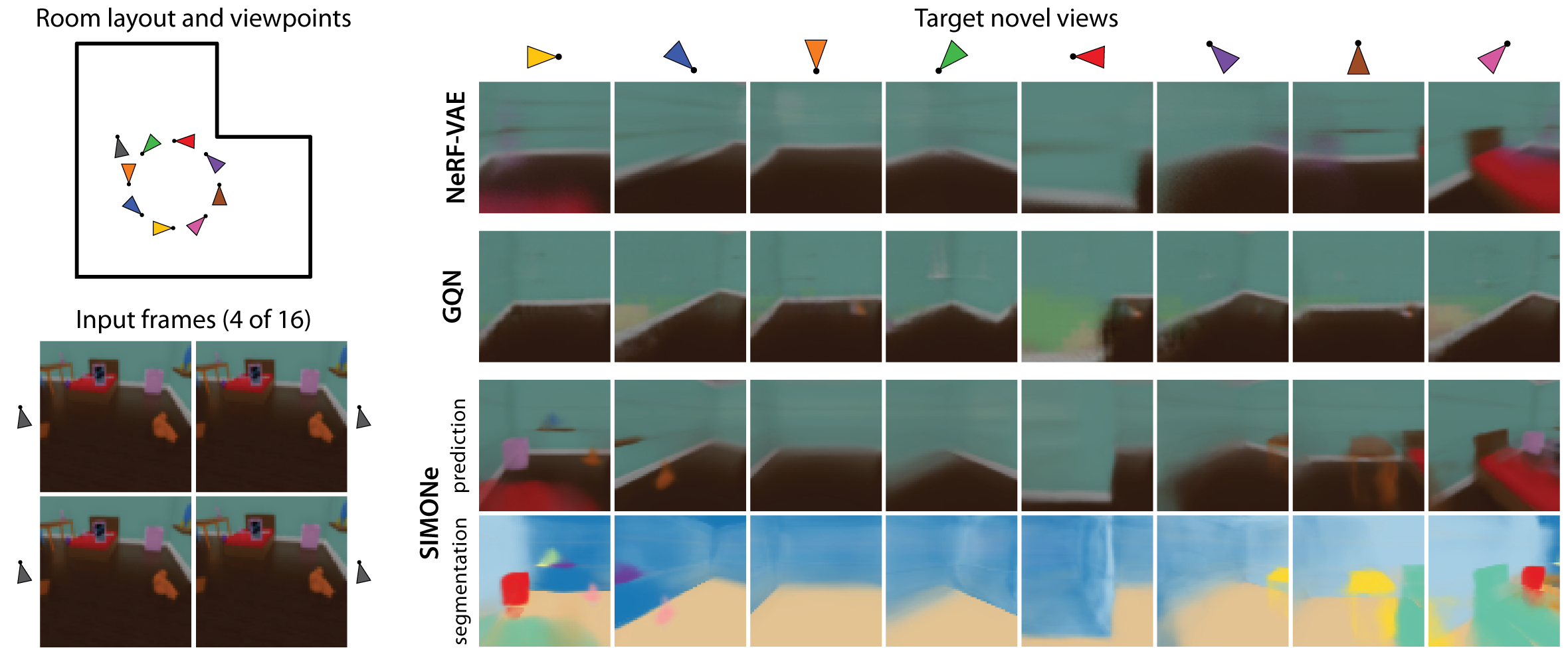}
    \caption{\textbf{Extended comparisons of scene representation and view synthesis capabilities} between \simonet-VS, NeRF-VAE, and GQN. Refer to Figures \ref{fig:view_interpolations}, \ref{fig:sa_view_interpolations_extra_1} for more details.}
\label{fig:sa_view_interpolations_extra_2}
\end{figure}

Figures~\ref{fig:sa_view_interpolations_extra_1}-\ref{fig:sa_view_interpolations_extra_2} show more examples of viewpoint traversals comparing \simonet-VS, NeRF-VAE and GQN. We execute the same traversal around the room (as in Figure~\ref{fig:view_interpolations}) given different input sequences (i.e. different rooms observed partially via different agent trajectories). Note that some context input sequences contain very little motion between frames (see last example in Figure~\ref{fig:sa_view_interpolations_extra_2}). The model still handles this well.

\begin{figure}[t]
    \centering
    \includegraphics[width=0.7\linewidth]{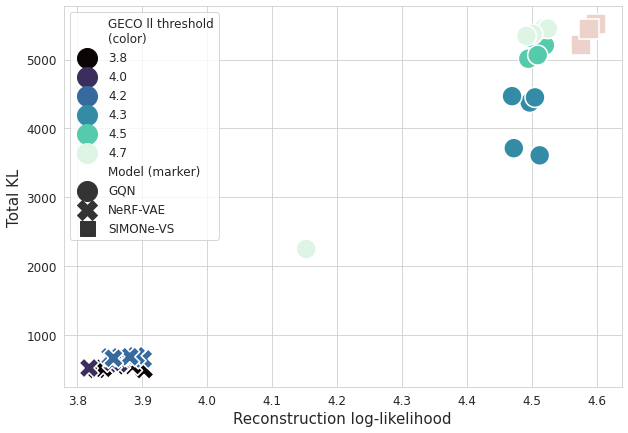}\\
    \caption{\textbf{Log-likelihood \& KL comparison across view-supervised models.} We show five independent runs of GQN and NeRF-VAE for three GECO log-likelihood thresholds each. For SIMONe-VS, we show three independent runs (trained as usual without GECO). Note that SIMONe-VS's KL shown here is the sum (not average) over K object latents.}
\label{fig:view_supervised_comparison_ll_kl}
\end{figure}

We also compare the view-supervised models based on their representation cost and reconstruction fidelity (rate and distortion respectively) in Figure~\ref{fig:view_supervised_comparison_ll_kl}. We first trained SIMONe-VS, and attempted to replicate its attained log-likelihood in the baseline models via constrained optimization (GECO). NeRF-VAE and GQN reconstructions saturate at the low end of the GECO thresholds we tried, implying bottlenecks that prevent the models from reaching higher log likelihoods. GQN does match SIMONe-VS in its KL representation cost (summed over K object latents in \simonet-VS's case). But as observed in the view synthesis images, GQN overfits to the context frames and is unable to interpolate between viewpoints successfully. 

This speaks to the effectiveness of \simonet{}'s object-centric latent structure. Generally speaking, we expect \simonet{} to achieve better reconstruction fidelity at the same total KL cost (or a similar reconstruction fidelity at lower KL cost) than a model which infers a single latent variable for the whole scene. 

\subsubsection{Instance segmentation}
\label{app:segmentation_comparison}

\begin{figure}[t]
    \centering
    \includegraphics[width=0.98\columnwidth]{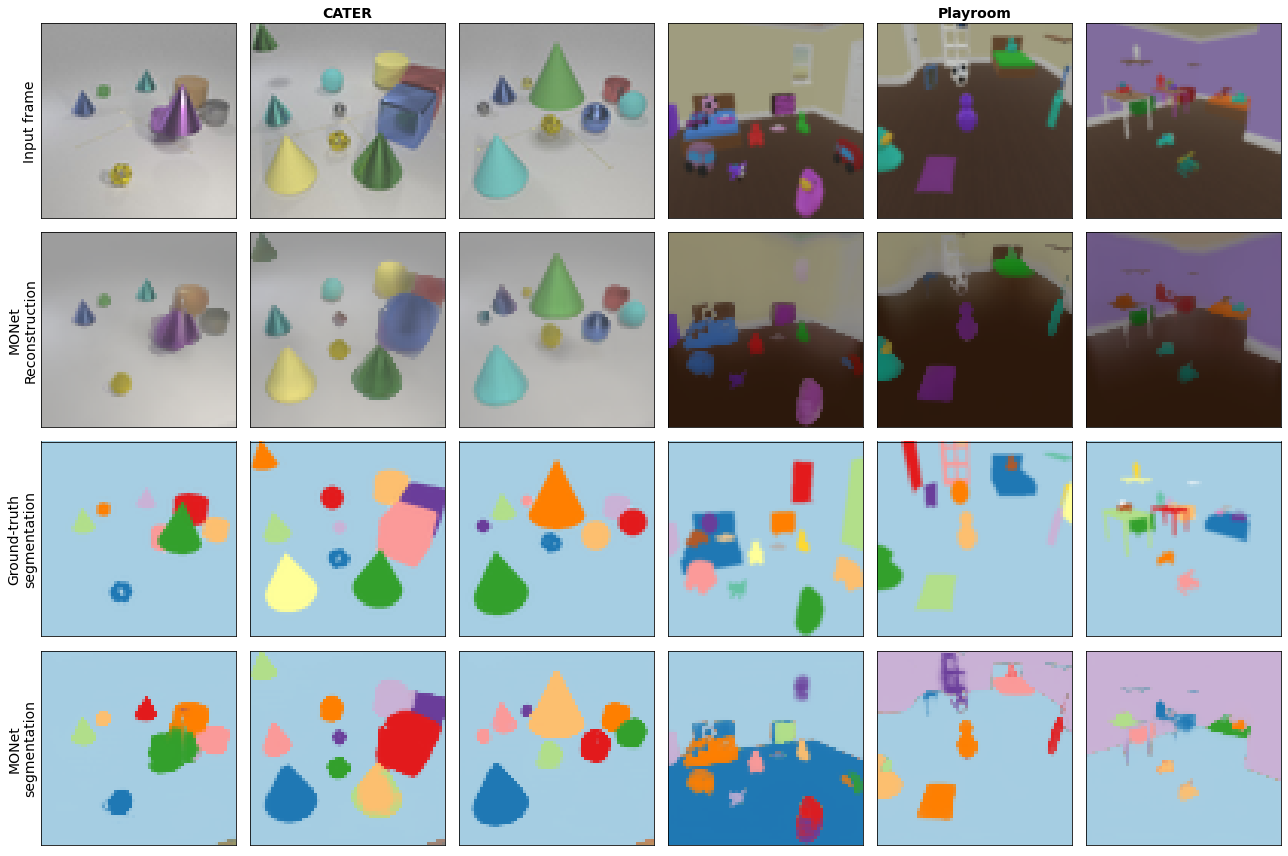}
    \caption{\textbf{Segmentations produced by MONet} on CATER and Playroom. Compare to Figure~\ref{fig:segmentations_qualitative} in main text.}
    \label{fig:sa_segmentations_monet}
\end{figure}

\begin{figure}[t]
    \centering
    \includegraphics[width=0.98\columnwidth]{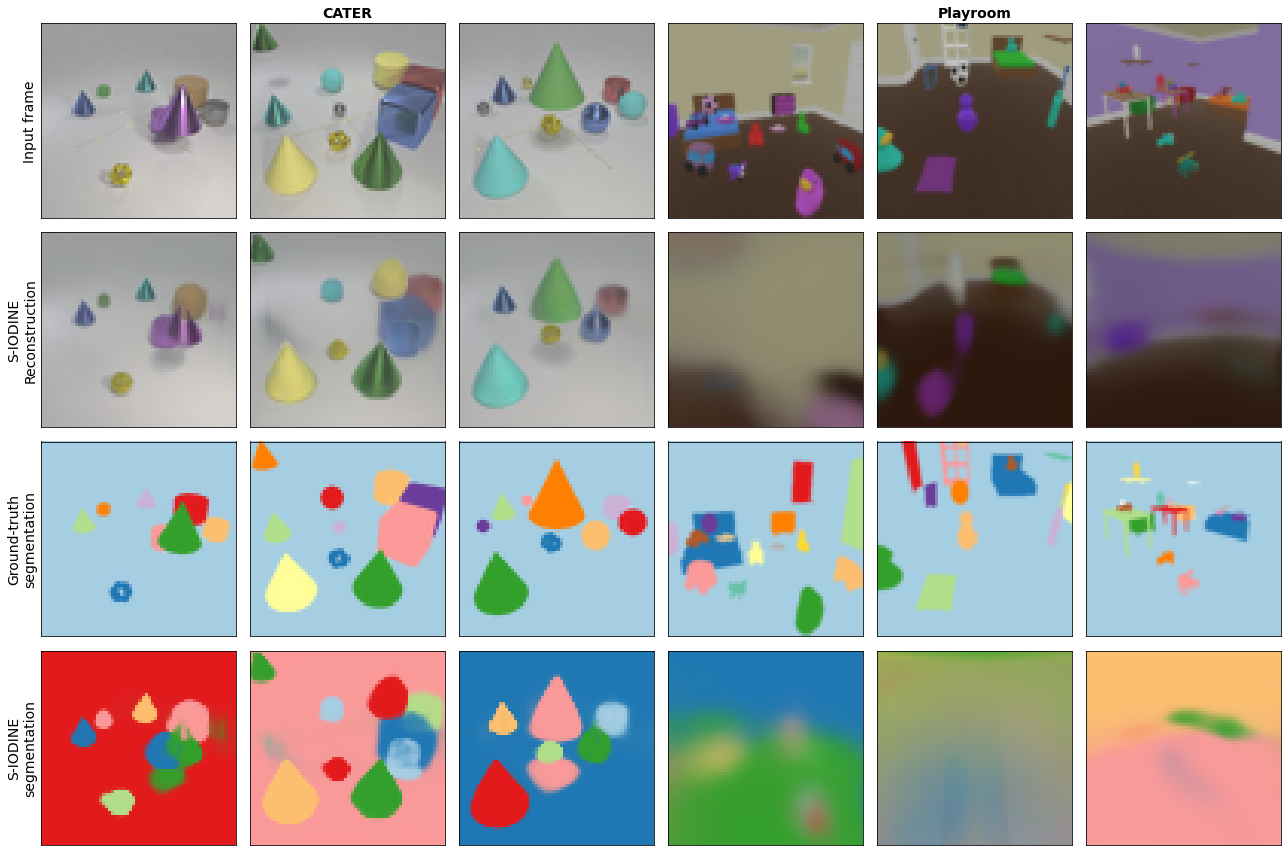}
    \caption{\textbf{Segmentations produced by S-IODINE} on CATER and Playroom. Compare to Figure~\ref{fig:segmentations_qualitative} in main text.}
    \label{fig:sa_segmentations_siodine}
\end{figure}

\begin{figure}[t]
    \centering
    \includegraphics[width=0.98\columnwidth]{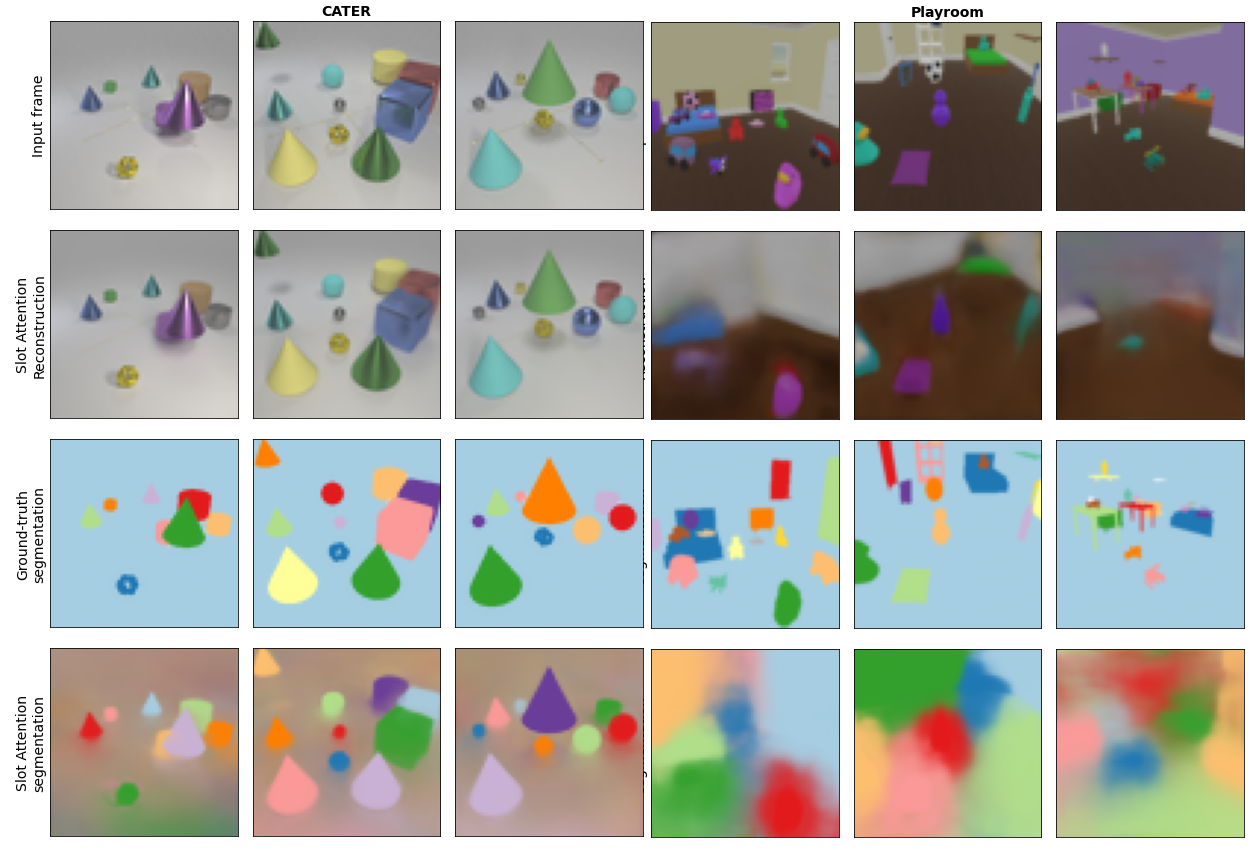}
    \caption{\textbf{Segmentations produced by Slot Attention} on CATER and Playroom. Compare to Figure~\ref{fig:segmentations_qualitative} in main text.}
    \label{fig:sa_segmentations_slotattention}
\end{figure}

Figures~\ref{fig:sa_segmentations_monet}-\ref{fig:sa_segmentations_slotattention} visualize how baseline models perform at segmenting CATER and Playroom scenes in comparison to the \simonet\ results we showed in Figure \ref{fig:segmentations_qualitative}. Note that for all segmentation maps visualized, we use soft (inferred) component mixture weights in combination with the color palette in Figure \ref{fig:color_palette}. This helps show each model's confidence in its segmentation. A blurry segmentation map (e.g. S-IODINE on Playroom data in Figure \ref{fig:sa_segmentations_siodine}) suggests higher-entropy component weights.

To elaborate on the qualitative differences between the models, note that \begin{enumerate*}[label=\textbf{(\arabic*})] \item MONet exhibits a tendency for clustering by color. On Playroom data, it consistently merges the brown base of each bed with the brown floor. It also groups toys/small objects by color (e.g. the blue mattress and blue-windowed bus in column 4 or the purple duck and purple cushion in column 5 of Figure~\ref{fig:sa_segmentations_monet}). On the other hand, SIMONe segments the full (two-colored) bed as one object consistently, and we don’t see it group the aforementioned toys by color.

\item When Slot Attention segments CATER, it ignores object shadows completely. It infers crisp shapes, showing a clear propensity to use each object’s uniform color and lack of texture. SIMONe on the other hand assigns every object’s shadows (up to three) in the corresponding object’s segment.
\end{enumerate*}

\subsubsection{Temporal abstraction and dynamics prediction}
\label{app:temporal_abstraction}

\begin{figure}[t]
    \centering
    \includegraphics[width=\linewidth]{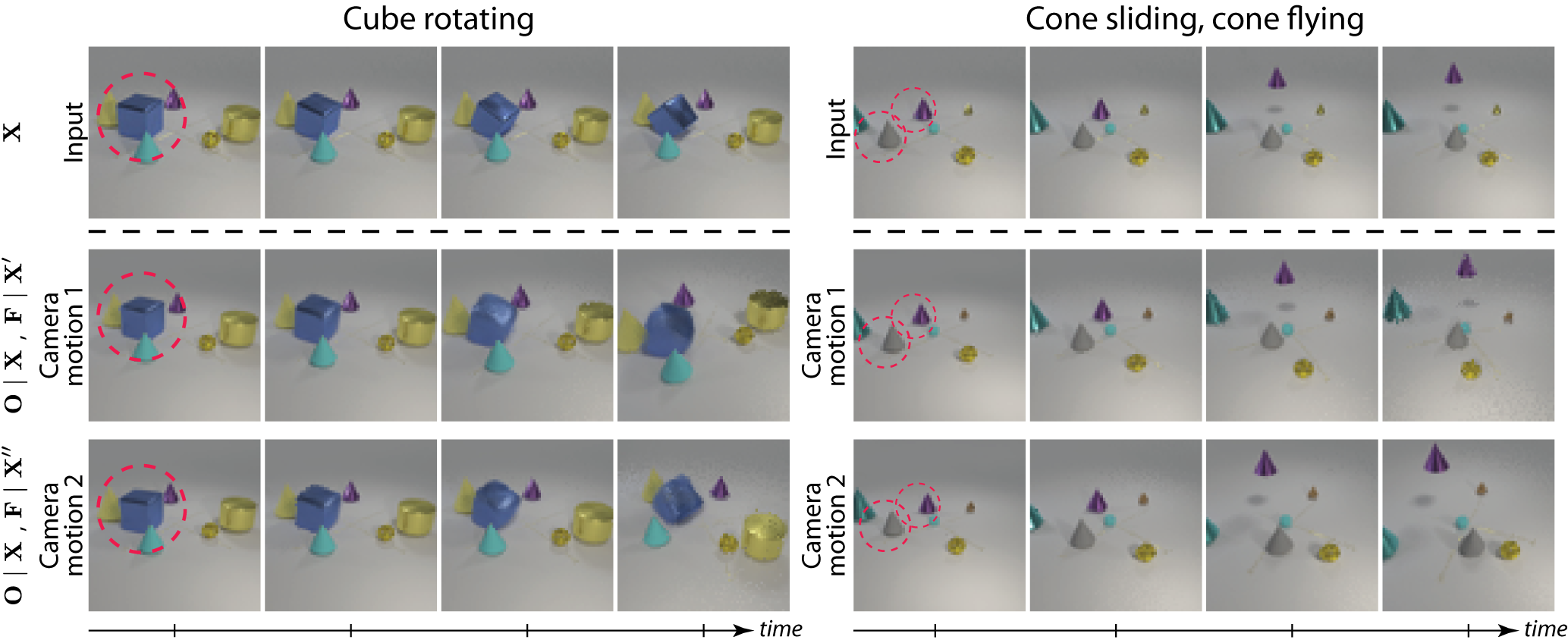}\\
    \vspace{5pt}
    \includegraphics[width=\linewidth]{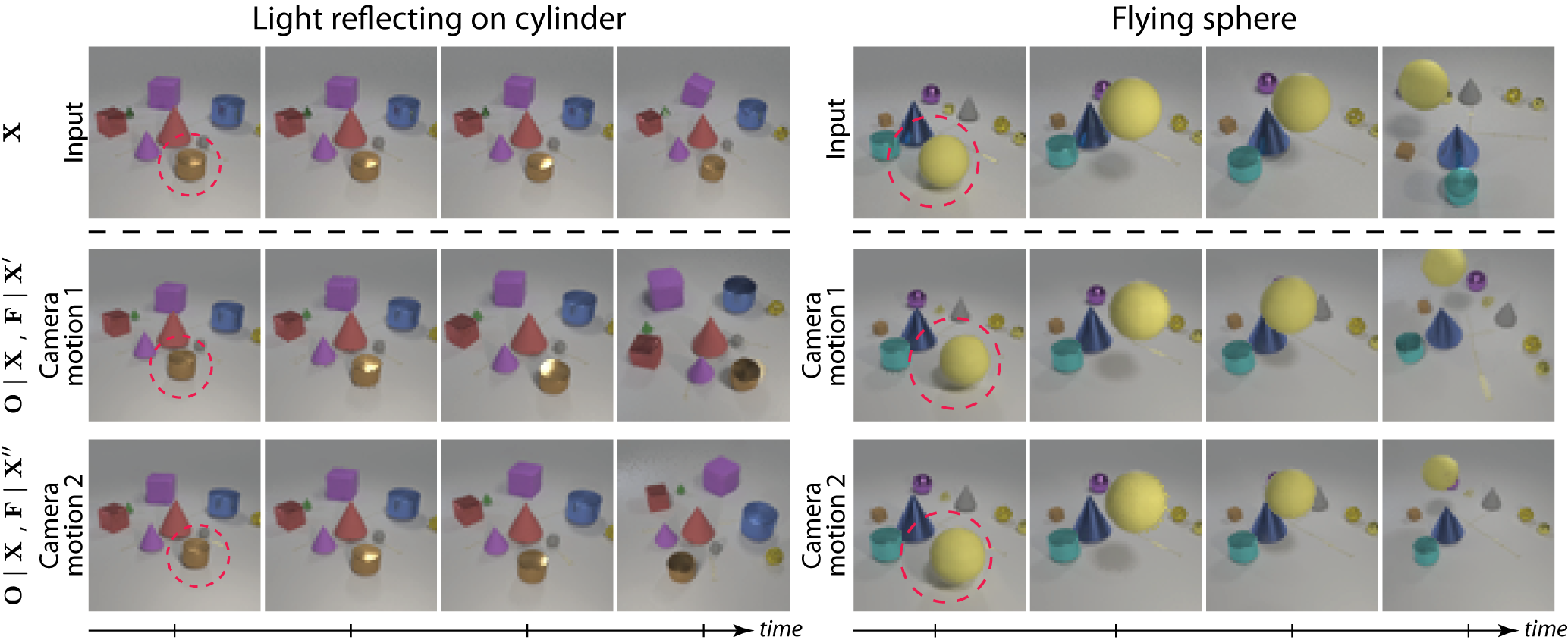}\\
    \caption{\textbf{Extended results showing separation of object trajectories from camera trajectories.} See Figure~\ref{fig:temporal_abstraction} for details.}
    \label{fig:sa_temporal_abstraction}
\end{figure}

We show more examples of the separation between object representations/trajectories and the camera's trajectory on CATER in Figure~\ref{fig:sa_temporal_abstraction}. In particular, we showcase other object dynamics present in the dataset (e.g. objects sliding on the floor or rotating), which \simonet{} also captures accurately. As before, the first row in each figure shows an input sequence $\X$. The other two rows reuse the object latents $\mathbf{O} | \X$ from the first sequence, but recompose them with frame latents from other (arbitrary) sequences: $\mathbf{F} | \X'$ and $\mathbf{F} | \X''$. We observe that the recomposed scenes are still composed of the same objects with their exact trajectories, while only the camera motion changes. This shows that object trajectories are represented invariantly of viewpoint (and vice versa).

\subsubsection{Frame latents}
\label{app:analysis}

\begin{figure}[t]
    \centering
    \includegraphics[width=0.9\columnwidth]{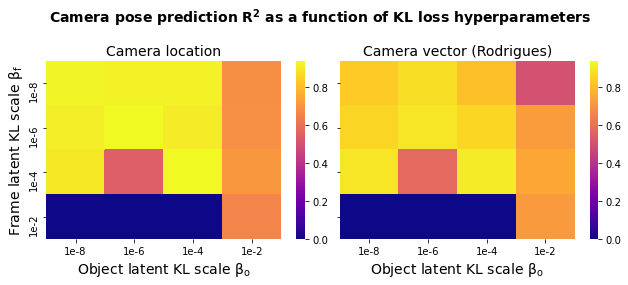}
    \caption{\textbf{Camera pose prediction given KL pressure hyperparameters.} Effect of $\beta_o$ and $\beta_f$ on the performance of predicting camera pose from frame latents ($R^2$ score, higher is better).}
    \label{fig:loss_hyperparams}
\end{figure}

To expand on our quantitative assessment of frame latents in Table~\ref{tab:predicting_camera_pose}, we look at the effect of hyperparameters $\beta_o$ and $\beta_f$ on the relationship between frame latents and camera pose (see Figure~\ref{fig:loss_hyperparams}). This relationship is one determinant of how "view-invariant" the object latents can possibly be.

\begin{figure}[h]
    \centering
    \includegraphics[width=\linewidth]{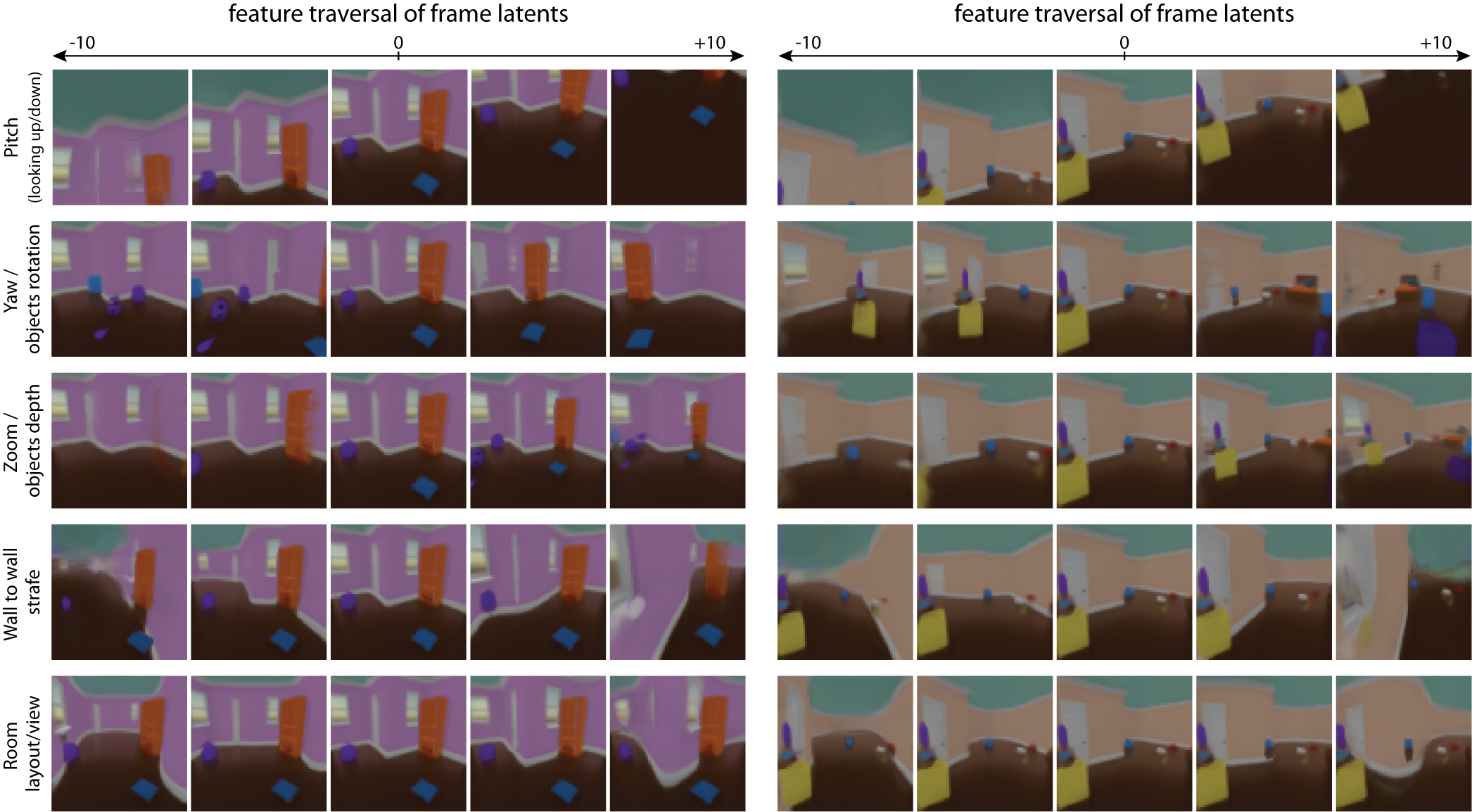}\\
    \vspace{5pt}
    \includegraphics[width=\linewidth]{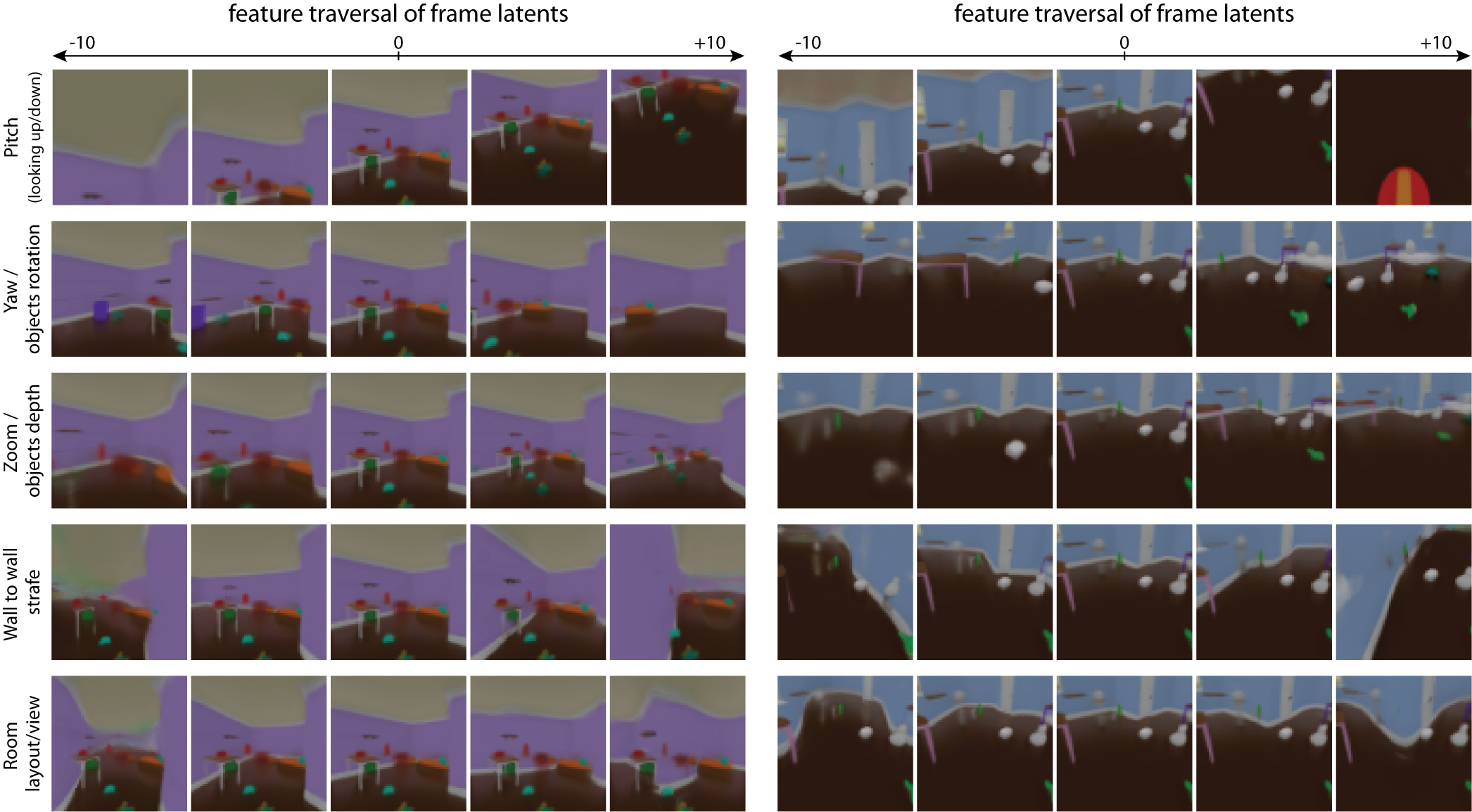}\\
    \caption{\textbf{Traversals of frame latent attributes.} We manipulate independent dimensions of a given frame latent. These should hypothetically control global/view attributes. (In contrast to Figure~\ref{fig:obj_latent_traversals}, all objects in the scene are affected when a frame latent attribute is manipulated). Each panel shows a different seed scene (visible in the middle column). The rows correspond to different latent attributes being manipulated (values are across the columns). We have labeled the rows with our interpretation for the effect of each latent attribute.}
    \label{fig:sa_frame_latent_traversals}
\end{figure}

Beyond their aggregate information content, we might also want that frame latents capture meaningful changes (e.g. in terms of view) per latent dimension. Figure \ref{fig:sa_frame_latent_traversals} shows the effect of individual frame latent attributes on Playroom. We selected the top ranking latents by marginal KL, and traverse them individually on several seed scenes. We indicate our interpretations of their behaviour in each row. Note that the latents controlling position in the room appear somewhat entangled. This may be a consequence of the policy used to collect our dataset; the agent's position is not arbitrary but influenced by the objects in the room.

\subsubsection{Composition of object latents from different scenes}

\begin{figure}[h]
    \centering
    \includegraphics[width=0.75\linewidth]{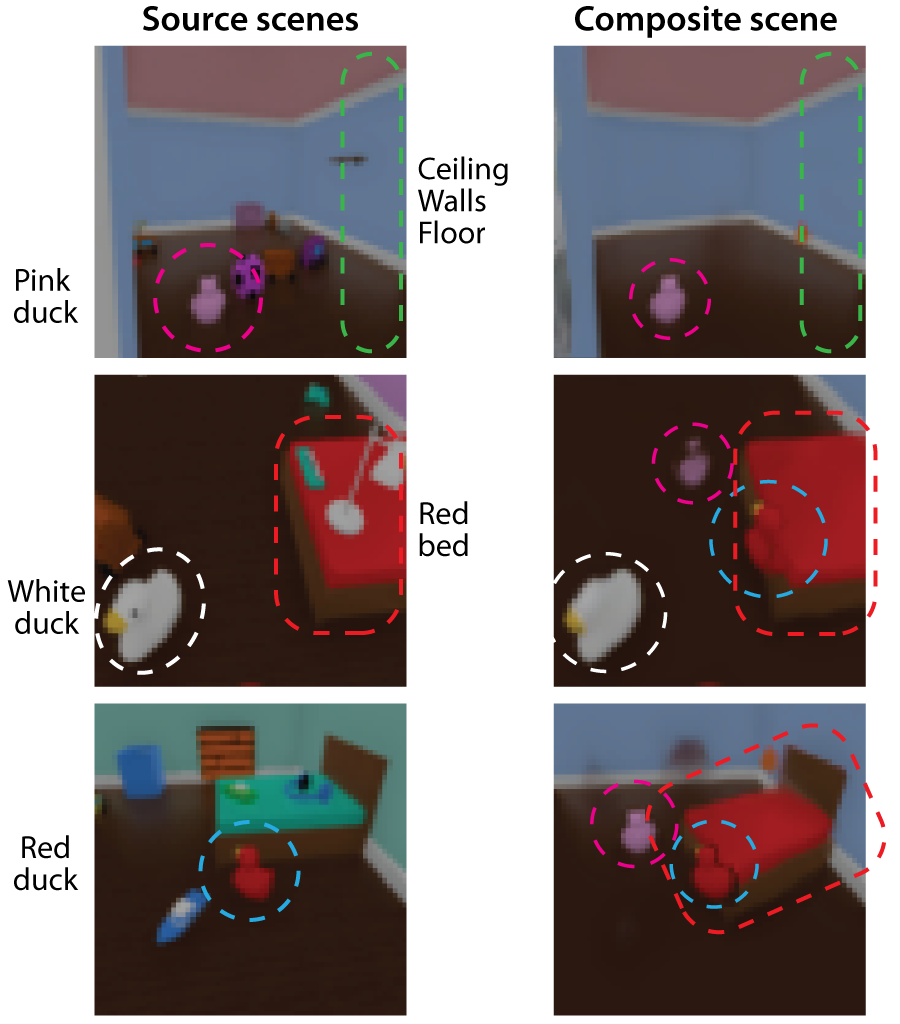}\\
    \caption{\textbf{Composition of object latents from different scenes.} \textbf{Left:} We take three different source scenes, and select different content from each (i.e. specific object latents, corresponding to the circled/labelled objects). \textbf{Right:} We then compose the selected contents into a novel set of objects (fewer than the usual $K=16$), which we can render from any viewpoint around the room (for this figure, we take the frame latent $\view_t$ from each source image on the left). Note that the model is able to cope with removing and adding objects, and renders them in a plausible fashion despite never being trained to do so.}
    \label{fig:sa_scene_composition}
\end{figure}

Given an object-centric representation, one should be able to manipulate scene contents and produce plausible compositional behaviour. This could involve removing or adding objects, swapping content between scenes, or varying the number of objects.
We present an early assessment of \simonet{}'s capabilities to perform such scene editing. We take a few different input scenes and compose (subsets of) their object latents into a novel composite scene, which can be then rendered from different points of view. See Figure~\ref{fig:sa_scene_composition} for an example.

\subsection{Wider Impact}
\label{app:wider_impact}  % !!! Has to be A.6 !!!

\simonet{} could benefit robotics and computer vision in multi-object scenes, whether indoors or on the street. There is some potential for misuse, especially in surveillance. While some prior work \cite{kosiorek2018sqair} has in fact used CCTV visuals of crowded scenes to demonstrate real use cases, we refrained from doing so.

\end{document}